\documentclass[sigconf]{acmart}

\setcopyright{none}
\renewcommand\footnotetextcopyrightpermission[1]{}

\usepackage{tikz}
\usepackage{amsmath}

\patchcmd{\maketitle}%
{\def\@makefnmark{\rlap{\@textsuperscript{\normalfont\@thefnmark\color{black}}}}}%
{\def\@makefnmark{\rlap{\@textsuperscript{\normalfont\color{black}\@thefnmark}}}}%
{}
{}
\makeatother 

\usepackage{filecontents}

\usepackage{hyperref}
\usepackage{url}
\usepackage{multirow}
\usepackage[utf8]{inputenc}
\usepackage[ruled, lined, longend, linesnumbered]{algorithm2e}

\usepackage{array}
\newcolumntype{L}[1]{>{\raggedright\let\newline\\\arraybackslash\hspace{0pt}}m{#1}}
\newcolumntype{C}[1]{>{\centering\let\newline\\\arraybackslash\hspace{0pt}}m{#1}}
\newcolumntype{R}[1]{>{\raggedleft\let\newline\\\arraybackslash\hspace{0pt}}m{#1}}

\usepackage{wrapfig}
\usepackage{colortbl}

\usepackage{subcaption}

\newcommand{\sys}{\texttt{LanFL}}

\begin{document}
	
	

    \title{Hierarchical Federated Learning through LAN-WAN Orchestration}
    
%
%
     \author{Jinliang Yuan}
     \affiliation{%
     \institution{Beijng University of Posts and Telecommunications}
     }
     
     \author{Mengwei Xu}
     \affiliation{%
     \institution{Beijng University of Posts and Telecommunications}
     }
     
     \author{Xiao Ma}
     \affiliation{%
     \institution{Beijng University of Posts and Telecommunications}
     }

     \author{Ao Zhou}
     \affiliation{%
     \institution{Beijng University of Posts and Telecommunications}
     }
     
     \author{Xuanzhe Liu}
     \affiliation{%
     \institution{Peking University}
     }
     
     \author{Shangguang Wang}
     \affiliation{%
     \institution{Beijng University of Posts and Telecommunications}
     }




	\maketitle


\subsection*{Abstract}

Federated learning (FL) was designed to enable mobile phones to collaboratively learn a global model without uploading their private data to a cloud server.
However, exiting FL protocols has a critical communication bottleneck in a federated network coupled with privacy concerns, usually powered by a wide-area network (WAN).
Such a WAN-driven FL design leads to significantly high cost and much slower model convergence.
In this work, we propose an efficient FL protocol, which involves a hierarchical aggregation mechanism in the local-area network (LAN) due to its abundant bandwidth and almost negligible monetary cost than WAN.
Our proposed FL can accelerate the learning process and reduce the monetary cost with frequent local aggregation in the same LAN and infrequent global aggregation on a cloud across WAN.
We further design a concrete FL platform, namely LanFL, that incorporates several key techniques to handle those challenges introduced by LAN: cloud-device aggregation architecture, intra-LAN peer-to-peer (p2p) topology generation, inter-LAN bandwidth capacity heterogeneity.
We evaluate LanFL on 2 typical Non-IID datasets, which reveals that LanFL can significantly accelerate FL training (1.5×–6.0×), save WAN traffic (18.3×–75.6×), and reduce monetary cost (3.8×–27.2×) while preserving the model accuracy.




\section{Introduction}
\label{sec:intro}

Federated learning (FL)~\cite{mcmahan2017communication} is an emerging machine learning (ML) paradigm that allows a ML model to be trained across a large number of clients collaboratively without requesting the data to be uploaded to a central server.
Due to its privacy-preserving nature and the ever-growing public concerns over user privacy (e.g., GDPR~\cite{gdpr}), FL has drawn tremendous attention in both academia and industry in recent years.
In this work, we focus on cross-device FL~\cite{kairouz2019advances}, where clients are resource-constrained end devices like smartphones.
It has a wide spectrum of use cases such as input method and item recommendation~\cite{bonawitz2019towards}.

\textbf{WAN-driven FL and its drawbacks.}
One critical characteristic of existing FL protocols, as compared to traditional distributed learning in datacenters, is that FL highly relies on wide-area network (WAN), e.g., the inter-city, inter-state, or even inter-country data transmission.
This is because, during a typical FL process, the central server iteratively collects model updates from a large number of geographically-distributed devices and dispatches aggregated results to them.
The aggregation typically repeats for 500--10,000 rounds before model convergence~\cite{kairouz2019advances}.
Such a WAN-driven design, however, leads to the following inevitable drawbacks.

$\bullet$ First, WAN is often highly constrained and unstable, e.g., as low as a few Mbps~\cite{wifi-state}.
As a result, the network has become a major bottleneck of the learning process and severely slows down the FL model convergence.
For example, Google reports that obtaining an RNN model for next-word prediction through FL takes 3000 rounds and 5 days~\cite{bonawitz2019towards}.
Existing solutions in reducing network communications cannot accelerate FL convergence due to straggler effect or compromised accuracy~\cite{yang2020heterogeneity}.

$\bullet$ Second, WAN is a precious, metered resource that adds billing cost to FL developers.
Our experiments show that executing one FL task till model convergence on AWS EC2 typically costs hundreds of dollars, among which 80\% is billed to network traffic.
Such a cost can be amplified by the need to periodically re-train the model and become prohibitive for small entities or individuals that want to practice in FL.

\textbf{Proposal for LAN-aware FL.}
In this work, we propose LAN-aware FL, a new FL paradigm that orchestrates with local-area network (LAN).
The rationales are that:
(1) Compared to WAN, LAN bandwidth resource is much more abundant, e.g., tens of Mbps as we will experimentally show in $\S$\ref{sec:bkgnd} (a nearly 10$\times$ gap).
This is because, during the WAN routing path, any hop can be the bandwidth bottleneck, e.g., the cloud service gateway can be a common one as it's shared by many tenants~\cite{fragkopoulos6reducing}.
In addition, LAN bandwidth is unmetered resource that does not add cost to cloud services.
Those advantages of LAN over WAN show the potential in accelerating FL convergence and reducing billing cost to deploy FL in real world.
(2) FL applications are commonly deployed on a large number of devices that are naturally organized into many LAN domains.
For each LAN domain such as campus and office building, there is often a substantial number of devices that are available for collaborative learning.
In fact, in $\S$\ref{sec:eval} we will experimentally show that 10 devices within the same LAN are enough to dramatically boost the FL performance.
(3) The functionality of direct data transmission via LAN is available on common mobile devices like Android~\cite{Android-P2P}, without relying on any third-party vendor support.

\begin{figure}
\centerline{\includegraphics[width=0.48\textwidth]{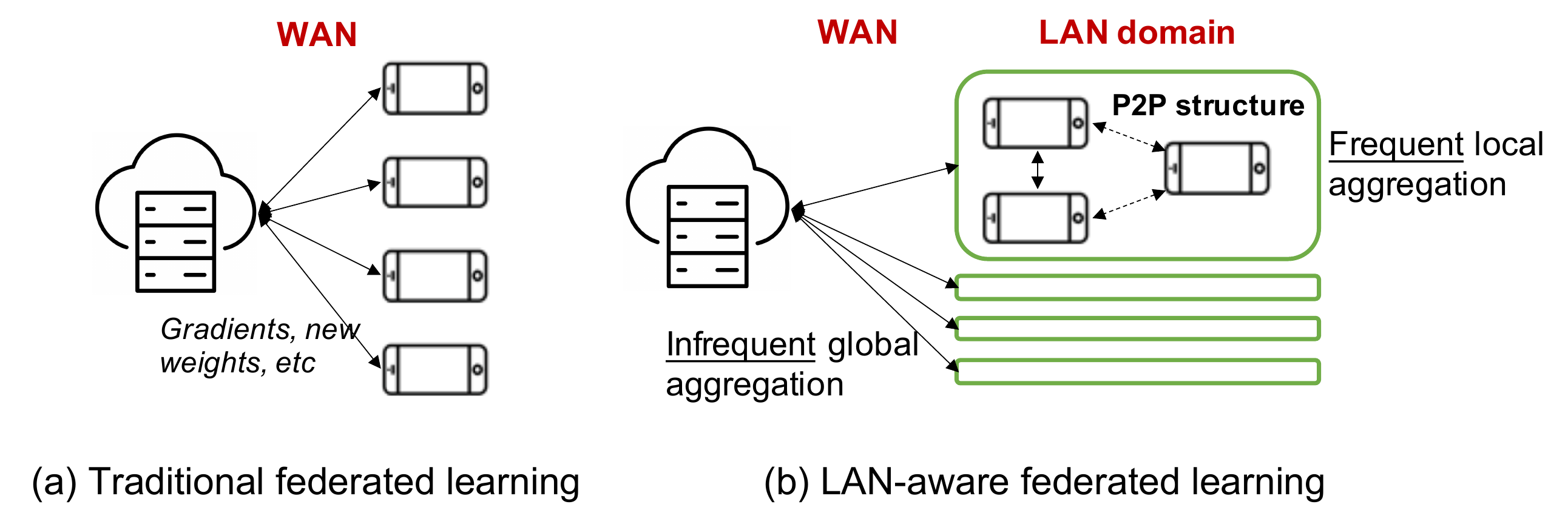}}
\vspace{-5pt}
\caption{A high-level comparison of traditional FL and LAN-aware FL (our proposal).}
\label{fig:overview}
\vspace{-5pt}
\end{figure}

While there're a few preliminary efforts in edge-assisted hierarchical FL~\cite{lim2020federated, liu2019edgeassist, abad2020hierarchical, briggs2020federated}, those approaches are built atop a cluster of edge servers which, however, are not publicly available yet and still suffer from WAN constraint~\cite{aws-local-zones,azure-edge-zone}.
Instead, we utilize LAN resources by organizing the devices in a peer-to-peer (P2P) mode without help from any edge server.
As shown in Figure~\ref{fig:overview}, the devices within the same LAN domain frequently exchange model updates (weights) and train a locally shared model.
Many LAN domains also upload their model updates to a central server but infrequently to collaboratively train a globally shared model.
Through this hierarchical design, WAN-based aggregation is much less needed and thus the learning can be accelerated.
Besides, throughout the learning process, no raw data leaves its host device so the user privacy is preserved as the original FL protocol.

\textbf{\sys: challenges and key designs.} Based on the above concept, we build \sys, the first LAN-aware FL framework to fully exploit the LAN bandwidth resources and relieve the learning process from WAN constraint.
In designing the hierarchical structure of \sys, we face the following challenges:
(1) \textit{How to coordinate the collaborative training across the devices within the same LAN domain, and across different domains?}
For example, we need to determine the proper frequency for intra-LAN and inter-LAN aggregation, considering their disparate bandwidth resources.
(2) \textit{How to organize the devices into a proper topology so that the LAN bandwidth can be efficiently utilized?}
A typical LAN domain comprises of many inter-connected access points (APs), and the devices connected to the same AP share the AP's bandwidth.
An improper topology may cause the end-to-end training performance being bottlenecked by one AP's capacity but leaving others under-utilized.
(3) \textit{How to coordinate the training pace across heterogeneous LAN domains with different network capacity?}
Such heterogeneity will lead to out-of-sync training pace of different LAN domains, and ultimately slows down the model convergence because the central server has to wait for slow LAN domains.

To address the above challenges, \sys{} incorporates several novel techniques.
First, \sys{} leverages the rich trade-off between training speed and model accuracy exposed by LAN domains, by separately tuning the parameters for intra-LAN and inter-LAN model aggregation.
Second, \sys{} adopts two widely-used topologies, i.e., parameter server~\cite{li2014scaling} and Ring-AllReduce~\cite{sergeev2018horovod} for intra-LAN collaborative training.
\sys{} judiciously constructs the network topology based on the device information profiled at runtime within each LAN domain.
Third, \sys{} exploits the shared LAN bandwidth and dynamically selects the participant device number to re-balance the training pace of heterogeneous LAN domains.

\textbf{Evaluation}
We evaluate \sys{} on two Non-IID datasets that are widely adopted in FL studies: FEMNIST~\cite{cohen2017emnist} and CelebA~\cite{liu2015deep}.
While preserving model accuracy, \sys{} is demonstrated to be very effective in accelerating model convergence (1.5$\times$--6.0$\times$), reducing WAN traffic (18.3$\times$--75.6$\times$), and saving monetary cost after deployment (3.8$\times$--27.2$\times$).
The effectiveness of \sys{}'s key design points are also evaluated separately.

\textbf{Major contributions} are summarized as following.

\begin{itemize}
\item We propose LAN-aware FL, a novel FL paradigm that can utilize the LAN bandwidth resource and thus relieve its reliance on constrained WAN.
\item We design a concrete LAN-aware FL simulation platform, namely \sys. It incorporates several novel techniques to address the unique challenges introduced by heterogeneous, hierarchical LAN design.
\item We comprehensively evaluate \sys{} atop 2 Non-IID datasets, which shows that \sys{} can significantly accelerate model convergence and reduce WAN traffic compared to traditional FL protocol.
\end{itemize}

\section{Background and Motivations}
\label{sec:bkgnd}

In this section, we first point out the fundamental drawbacks of WAN-driven design in FL and the potential benefits of switching to LAN ($\S$\ref{sec:bkgnd-LAN}).
We then identify the key challenges in designing an efficient, LAN-aware FL protocol ($\S$\ref{sec:bkgnd-challenges}).

\subsection{From WAN to LAN:  Rationales}\label{sec:bkgnd-LAN}

\textbf{Federated learning and its WAN}
Traditional federated learning protocols heavily rely on wide-area network (WAN) because, as exemplified by \textit{FedAvg} in Algorithm 1,
the central server needs to iteratively collect model updates from a large number of remote, decentralized devices and dispatch the aggregated results back to them.
In addition, it often takes thousands of rounds before model convergence.
We identify the following drawbacks of such WAN-driven learning process.

$\bullet$
\textit{First, WAN is known to be highly constrained and unstable~\cite{wifi-state}, which can severely slow down the convergence of FL training}, e.g., 3,000 rounds and 5 days for an RNN model reported by Google~\cite{bonawitz2019towards}.
In fact, network transmission has become a bottleneck in the FL process~\cite{li2018federated, smith2017cocoa}.
To mitigate the high network overhead, typical FL algorithms like \textit{FedAvg} resort to increase local training epochs to re-balance the computation-communication time. Such an approach, however, increases the straggler affect among distributed models and the aggregation becomes less effective given the typically non-convex NN models and Non-IID datasets cross devices.
Other techniques like gradients compression \cite{caldas2018expanding, wang2018atomo, tang2018communication} also save network bandwidth per round but lead to accuracy degradation.
As a result, the time to converge even increases~\cite{yang2020heterogeneity}.

$\bullet$
\textit{Second, WAN usage incurs a high monetary cost to FL practitioners.}
Nowadays, major cloud service providers like AWS and Azure support charging the network cost on demand.
Since federated learning requires very little computational and memory resources (only for weights aggregation), the network cost often dominates the total monetary cost, e.g., more than 80\% as we estimate in $\S$\ref{sec:eval}.
In reality, the cost is amplified by the demand to periodically update the model, e.g., to adapt to the evolving data distribution (input method) or new advanced model structures~\cite{xu2018deeptype}.
As a result, it becomes prohibitive for many small entities or individuals that want to practice on federated learning.

\textbf{The opportunity of LAN-aware FL}
To reduce its reliance on WAN, it's intuitive to orchestrate the FL process with local-area-network (LAN).
Its feasibility is recognized by three facts:
(1) LAN resource is much more abundant than WAN, as the data packets do not go through the backbone network.
For example, typical wireless routers can transmit data at hundreds of Mbps.
(2) Nowadays internet infrastructure is organized across many LANs, where each LAN may cover a large number of end devices such as campus, office building, hospital, etc.
From the FL developer (or app developer) perspective, the users that can participate in FL are also clustered by LANs.
(3) The end devices within the same LAN can discover and communicate with each other in a P2P model without experiencing WAN.
For example, the Android system provides \texttt{Wi-Fi Direct} API~\cite{Android-P2P} that enables service discovery, broadcast, and encrypted transmission.
Noting that WiFi is the major internet access method for mobile devices in FL because it's unmetered~\cite{bonawitz2019towards}.

From the whole internet perspective, WAN is a limited, precious resource that shall be preferentially provisioned to user-interactive applications like video streaming other than FL.
Our proposal of LAN-aware FL inherits the spirit of edge computing~\cite{shi2016edge} and can help relieve the fast-growing tension on the global internet backbones.

\subsection{Challenges}\label{sec:bkgnd-challenges}

Given the inherent advantages of LAN over WAN in decentralized training, it's not easy to efficiently utilize LAN resources while preserving the original dignity of FL performance, e.g., converged model accuracy.
More specifically, we identify following unique challenges introduced by LAN.

$\bullet$ \textbf{Hierarchical design}
Since each LAN domain usually has limited number of devices, which is not enough to train high-quality ML models, a practical LAN-aware FL design still needs WAN to connect many LAN domains.
It leads us to a hierarchical design, where models will be aggregated both locally (intra-LAN) and globally (inter-LANs).
Given the disparate behaviors of WAN and LAN bandwidth resources, however, it becomes difficult to orchestrate them in a harmonious way.
$\S$\ref{subsec:LAN-design} will show how we address this challenge by tuning the key training parameters for intra-LAN and inter-LAN collaborative learning separately.

\begin{wrapfigure}{r}{0.28\textwidth}
	\centering
	\includegraphics[width=.28\textwidth]{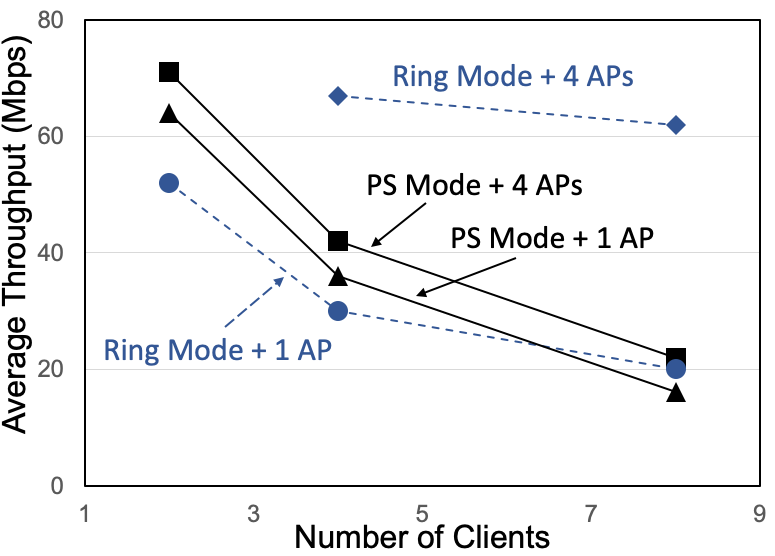}
	\caption{A case study of P2P WiFi throughput in a campus LAN. ``PS mode'': parameter server; ``Ring mode'': Ring-AllReduce; AP: access point.}
	\label{fig:campus-lan}
\end{wrapfigure}

$\bullet$ \textbf{LAN bandwidth sharing}
Existing FL platforms and literature assume static or pre-assigned WAN bandwidths to each device without considering their interference.
However, the intra-LAN network throughput highly depends on the number of devices that are transmitting data simultaneously.
It opens extra trade-off between the parallel training devices and data communication speed.
Such trade-off is unexplored yet critical to system performance.

Figure~\ref{fig:campus-lan} shows a case study of P2P WiFi throughput in a campus LAN.
Here, PS/Ring modes represent different network topology structures about how devices are inter-connected, where the former imitates parameter server~\cite{li2014scaling} (1 server) and the latter imitates Ring-AllReduce~\cite{sergeev2018horovod}.
Both of them are classical topologies commonly used in datacenters.
$\S$\ref{subsec:lan-aware-cc} will further elaborate the difference between those two topologies.
``AP'' is the number of access points (routers) that connect to the devices.
For example, 8 devices and 4 APs mean that each 2 devices are connected to one AP while 4 APs are within the same LAN.
The figure shows that:
(1) Under PS mode, the throughput decreases almost linearly with increasing number of devices, as the server side becomes the bottleneck of transmission path.
(2) When devices are distributed under many APs' coverage, Ring mode can significantly increase the throughput because it can efficiently utilize the available bandwidth of each AP.
Those findings provide crucial insights to our algorithm design in $\S$\ref{subsec:lan-aware-cc}.

$\bullet$ \textbf{LAN Heterogeneity}
To obtain a high-accuracy, unbiased model, many LAN domains need to collaborate with others.
This is because one LAN domain typically has only a small number of available devices that are not enough to train the model as we will experimentally show in $\S$\ref{eval:number}.
However, cross-LAN collaboration brings in the LAN heterogeneity, i.e., the diverse LAN bandwidth that can be utilized for model parameters exchange.
This is because one LAN domain typically has only a small number of available devices that are not enough to train the model.
However, cross-LAN collaboration brings in the LAN heterogeneity, i.e., the diverse LAN bandwidth capacity that can be utilized for model parameters exchange.
The heterogeneity can be caused by many factors, including hardware capacity of wireless access point, the sharing condition by other devices, etc.
A straightforward approach is to synchronize on cloud and wait for the slow LAN domains to complete the same number of aggregations as others.
This approach, however, slows down the model convergence.
$\S$\ref{subsec:design-heter} shows how we deal with such heterogeneity.

\section{\sys{}: an LAN-aware FL Design}
\label{sec:design}

\subsection{Problem Definition}


\begin{table}[]
\footnotesize
\begin{tabular}{|L{1.5cm}|L{6.2cm}|}
\hline
\textbf{Notation} & \textbf{Descriptions} \\ \hline \hline
cloud round  & Round number of cloud aggregation ($RW$) \\ \hline
device round  & Round number of device aggregation in each LAN ($RL$) \\ \hline
$BW$ & WAN bandwidth between devices and cloud \\ \hline
$BL$ & LAN bandwidth between devices (P2P) \\ \hline
$\omega$ & Global model stored in cloud \\ \hline
$G$ & A graph represents LAN domain collective communication topology (details in $\S$\ref{subsec:lan-aware-cc}) \\ \hline
$NL$ & Total number of LAN domain \\ \hline
$CG$ & Device group in each LAN domain \\ \hline
$CT$ & The group of selected training devices in LAN domain \\ \hline
$NC$ & Total number of device in each LAN domain\\ \hline
$N$ & Total number of device in WAN \\ \hline
$E$ & Local epochs of device training \\ \hline
\end{tabular}
\caption{Symbols and descrptions}
\label{tab:term}
\end{table} 
\begin{algorithm}
\caption{Federated Averaging (FedAvg)}\label{algo:FedAvg}

\small
\SetKwInOut{Input}{input}
\SetKwInOut{Output}{output}
\SetKw{KwBy}{by}

\Input{$N$, $RW$, $E$, $\omega_0$
}
\Output{the global model $\omega$}

\SetKwProg{Fn}{Function}{:}{}

\Fn{FedAvg()}{
	\For{$t\gets0$ \KwTo $RW$}{
        $S_t \gets$ select $N_s(N_s \in N)$ training devices randomly \\
        \For{each device $C_k(k \in N_s)$ in $S_t$ \textbf{in parallel}}{
            push $\omega_t$ to $C_k$ \textbf{across WAN} \\
            $\omega_t^k \gets$ train $\omega_t$ on local training dataset of $C_k$ for $E$ epoches \\
		    push  $\omega_t^k$ to the cloud \textbf{across WAN}\\
        }
        $\omega_{t+1} \gets$ $\frac{1}{N_s}$$\sum_{k=1}^{N_s}$$\omega_t^k$
	}
}
\end{algorithm}

FL protocols are designed to handle multiple devices collecting data and a central server coordinating the global learning objective across the network \cite{li2018federated}.
Here the network refers specifically to the WAN among the devices and the cloud. The objective can be formulated as:
\begin{equation}\label{eq:WAN-FL}
    \setlength{\abovedisplayskip}{3pt}
    \setlength{\belowdisplayskip}{3pt}
    \min_{\omega} f(\omega) = \sum_{k=1}^{N}\frac{m_kF_k(\omega)}{n},
\end{equation}
where $m_k$ is the number of training samples in device $k$, $n$ is total number of training samples in all devices and $F_k(\omega)$ is the loss function of device $k$.
In existing FL protocols, e.g., \textit{FedAvg} as described in Algorithm~\ref{algo:FedAvg}, the aggregation is performed on cloud with the $F_k(\omega)$ sent by each device $k$ across WAN.
However, a critical fact is ignored that most devices are distributed across different LANs.
Those devices in the same LAN can adopt P2P communication mechanism to aggregate models, which is much faster than the cloud aggregation across WAN.
Intuitively, the objective of our proposed \sys{} framework is to accelerate the WAN-driven FL protocol while ensuring model accuracy. 
Our \sys{} introduces a LAN domain aggregation mechanism to achieve our goal, which can be formulated as:
\begin{equation}\label{eq:lan-fl-cloud}
    \setlength{\abovedisplayskip}{3pt}
    \setlength{\belowdisplayskip}{3pt}
    \min_{\omega} f(\omega) = \sum_{i=1}^{NL}\frac{l_i\mathbf{F}_i(\omega)}{n},
\end{equation}
where $l_i$ is the number of total training samples in each LAN $i$, and $\mathbf{F}_i(\omega)$ is the loss function of LAN $i$. 
The loss function of LAN $\mathbf{F}_i(\omega)$ can be formulated as:
\begin{equation}\label{eq:lan-fl-device}
    \setlength{\abovedisplayskip}{3pt}
    \setlength{\belowdisplayskip}{3pt}
    \mathbf{F}_i(\omega) = \sum_{j=1}^{NC_i}\frac{m_jF_j(\omega)}{l_i},
\end{equation}
where $m_j$ denotes the number of  training samples on device $j$ in LAN $i$, $NC_i$ denotes the number of devices in LAN $i$, and $F_j(\omega)$ is the loss function of device $j$ in LAN $i$.
The $f(\omega)$ in Eq. \ref{eq:lan-fl-cloud} and $\mathbf{F}_i(\omega)$ in Eq. \ref{eq:lan-fl-device} are respectively calculated on cloud (across WAN) and on aggregating device in each LAN $i$ (across LAN).


\subsection{\sys{} Workflow}
\label{subsec:workflow}


\let\oldnl\nl
\newcommand{\nonl}{\renewcommand{\nl}{\let\nl\oldnl}}

\begin{algorithm}
\caption{Our proposed \sys{} framework}\label{algo:workflow}

\small
\SetKwInOut{Input}{input}
\SetKwInOut{Output}{output}
\SetKw{KwBy}{by}
\Input{the init model $\omega_0$,
    the set of LAN domain device group $\mathbb{CG}$,
    round number $RL$, round number $RW$,
    local epochs $E$
}
\Output{the global model $\omega$}

\SetKwProg{Fn}{Function}{:}{}

\Fn(\tcp*[h]{run on cloud}){Cloud\_Operation()}{
	\For{$k\gets0$ \KwTo $RW$}{
        $\mathbb{CG}_s \gets$ select subset of LAN domain participating in FL \textbf{randomly} \\
        \For{each $CG_i$ in $\mathbb{CG}_s$ \textbf{in parallel}} {
            $\omega_k \gets$ invoke $CG_i.$\textit{LAN\_Orchestrating}($\omega_k$) \\
	    }
		$\omega_{k+1} \gets$ $\sum_{NL}\omega_k$ \\
        update $\mathbb{CG}$ with latest information \\
	}
}

\Fn(\tcp*[h]{run on cloud}){LAN\_Orchestrating($\omega_k$)}{
    \For{$j\gets0$ \KwTo $RL$}{
        $CT_j \gets$ select training devices from $CG_i$ \textbf{randomly} \\
        $G_j \gets$ get PS or Ring topology with $CT_j$ \\
        $A_j \gets$ specify the aggregating device in $CT_j$ by $G_j$ \\
        \For{each device $C$ in $CT_j$ \textbf{in parallel}}{
			push $\omega_k$ to $C$ \textbf{across WAN} \\
			$\omega_k \gets$ invoke $C.$\textit{Device\_training}($\omega_k$, $A_j$) \\
		}
        $\omega_k \gets$ invoke $A_j.$\textit{Device\_aggregating}($\omega_k$, $RL$, $G_j$) \\
    }
}

\Fn(\tcp*[h]{run on device}){Device\_training($\omega_k$, $A_j$)}{
    update $\omega_k$ on local training dataset for $E$ epochs\\
    push  $\omega_{k}$ to aggregating device $A_j$ \textbf{across LAN} \\
}

\Fn(\tcp*[h]{run on device}){Device\_aggregating($\omega_k$, $RL$, $G_j$)}{
    $\omega_{k} \gets$ $\sum_{NC_s}\omega_k$, where $NC_s$ denotes the size of  $CT_j$\\
    \If{$j \textless RL$}{
        push $\omega_{k}$ to training devices specified in $G_j$ \textbf{across LAN} \\
    }
    \If{$j == RL$}{
        push $\omega_{k}$  to the cloud \textbf{across WAN}\\
    }
}
\end{algorithm}

The pseudo code in Algorithm \ref{algo:workflow} shows the workflow of \sys{}.
Overall, \sys{} also adopts a C/S architecture, where a central server maintains and keeps advancing a global model $\omega$, but the ``client'' refers to not only one device but an LAN domain comprising many devices connected to each other through P2P mechanism.
\sys{} follows \textit{FedAvg} to aggregate the model weights updated by many local training, but such aggregation happens on both devices (intra-LAN) and cloud (inter-LAN).
We next elaborate the jobs of cloud and devices.

\textbf{Device side}
The devices in \sys{} are split into two types: training devices and aggregating devices.
As a training device, it is responsible for updating local model and pushing them to the aggregating device specified by $G$, which is received from the cloud (line 22-23).
As an aggregating device, it frequently aggregates models received from training devices (line 25).
Then, according to the device round $RL$, it decides whether to send the result to the cloud for global aggregation or to the training devices specified by the $G$ to continue the local update (line 26-31).

\textbf{Cloud side} is mainly responsible for following three tasks.

First, it maintains a list of global information, including the global model ($\omega$), LAN domain information (e.g., $\mathbb{CG}$) and on-device training parameters (e.g., $E$).
The LAN domain information is collected from the devices participating in \sys. It will be periodically updated at the end of each cloud round $RW$ (line 8).
Among those factors, prior studies and our experiments show that $RL$ and $E$ play critical roles in the end-to-end learning performance~\cite{li2018federated, mcmahan2017communication}.
Thus, we introduce a LAN domain design in \sys{} to tune those parameters in $\S$\ref{subsec:LAN-design}.

Second, the cloud first selects the LAN domains participating in FL training at each cloud round $RW$, and then orchestrates their parallel operations (line 3-6). It also orchestrates the training devices selection in each LAN domain for parallel on-device training (line 15-18).
To avoid introducing additional bias,  \sys{} adopts a random strategy when selecting LAN domain and training devices (line 3, 12). 
In addition, we design a LAN-aware collective communication topology $G$ with the selected training devices and their meta information to efficiently aggregate models in each LAN domain (line 13).
This topology specifies the aggregating devices and orchestrates the device aggregation in each LAN domain (line 14, 19).
The detailed design is elaborated in $\S$\ref{subsec:lan-aware-cc}.

Third, the cloud needs to iteratively update the global model by aggregating the received models.
We follow the \textit{FedAvg} to update the global model (line 7).
Noting that \sys{} only needs one aggregating device in each LAN domain, for both PS mode and Ring mode, to transmit its aggregated model to the cloud (line 30). 
This design can save a lot of WAN traffic in \sys{}.

\subsection{LAN Domain Design}
\label{subsec:LAN-design}

Many prior studies have shown the importance of properly tuning the hyperparameters of \textit{FedAvg} due to the Non-IID settings in FL \cite{li2018federated, mcmahan2017communication}.
In particular,  the local epoch number $E$ exposes rich trade-off between training time and model accuracy.
On one hand, a larger number of  $E$ allows for more local computation on devices and potentially reduces cloud round across WAN, which can greatly accelerate the overall convergence process.
On other hand, with Non-IID datasets distributed cross devices, a larger number of $E$ may lead each device towards the optima of its local objective as opposed to the global objective, which potentially hurts convergence accuracy~\cite{li2018federated, bonawitz2019towards}.
In essence, such a contradiction is due to the slow WAN between the cloud and devices.

\begin{table}[H]
\centerline{\includegraphics[width=0.48\textwidth]{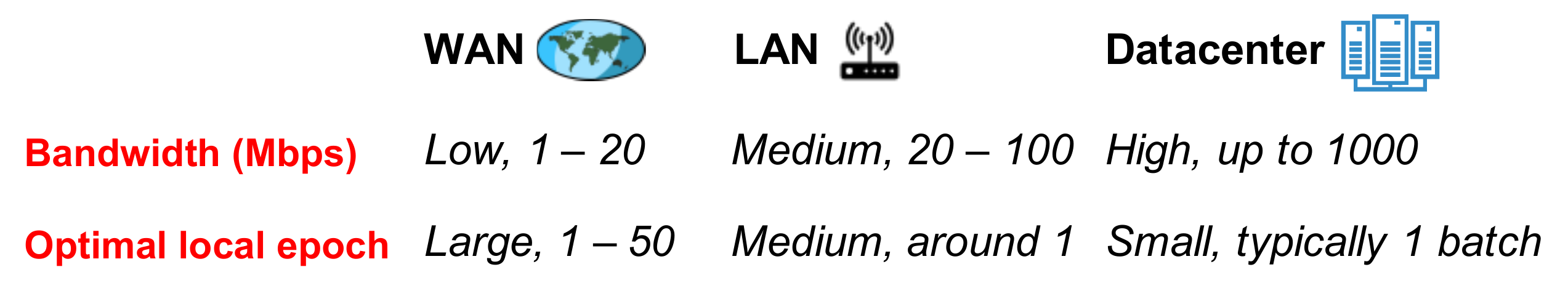}}
\caption{A comparison of WAN, LAN, and datacenters in consideration of their bandwidth capacity and optimal local epoch in distributed learning.}
\label{tab:wan-lan-dc}
\vspace{-10pt}
\end{table}

To some extent, \sys{}'s LAN domain is similar to a distributed system in datacenter, where many machines are inter-connected through switches.
As summarized in Table~\ref{tab:wan-lan-dc}, regarding the network bandwidth capacity, however, LAN is somewhere between the WAN and datacenter.
It indicates a new spectrum of parameter tuning different from WAN-driven FL (typically very large $E$) and datacenter (typically as small as one batch per aggregation).

\sys{} introduces a LAN domain module, within which the models are aggregated more frequently: with fewer local epochs $E$ but more device round $RL$.
The module is used to aggregate those device models in each LAN domain with $RL$ device rounds. In each device round $RL$, those selected training devices update local model with $E$ epochs in parallel.
We explain in two aspects why this design can speed up FL while ensuring accuracy.
First, a smaller $E$ can ensure the accuracy for the Non-IID datasets on devices in each LAN domain. It doesn't increase the communication cost in LAN too much because the communication speed in LAN is much faster than that in WAN.
Second, a larger $RL$ can reduce the number of aggregation in cloud across WAN. It doesn't significantly reduce the global model accuracy because the cloud aggregates the entire LAN domain model, not the more biased device model with more Non-IID datasets.
\sys{} currently relies on heuristics to tune the above training parameters.
In the future, \sys{} may apply smarter approaches~\cite{bonawitz2019federated,kairouz2019advances}, e.g., learning-based techniques, to relieve developers' burdens.

\subsection{LAN-aware Collective Communication}
\label{subsec:lan-aware-cc}

\begin{figure}
\centerline{\includegraphics[width=0.48\textwidth]{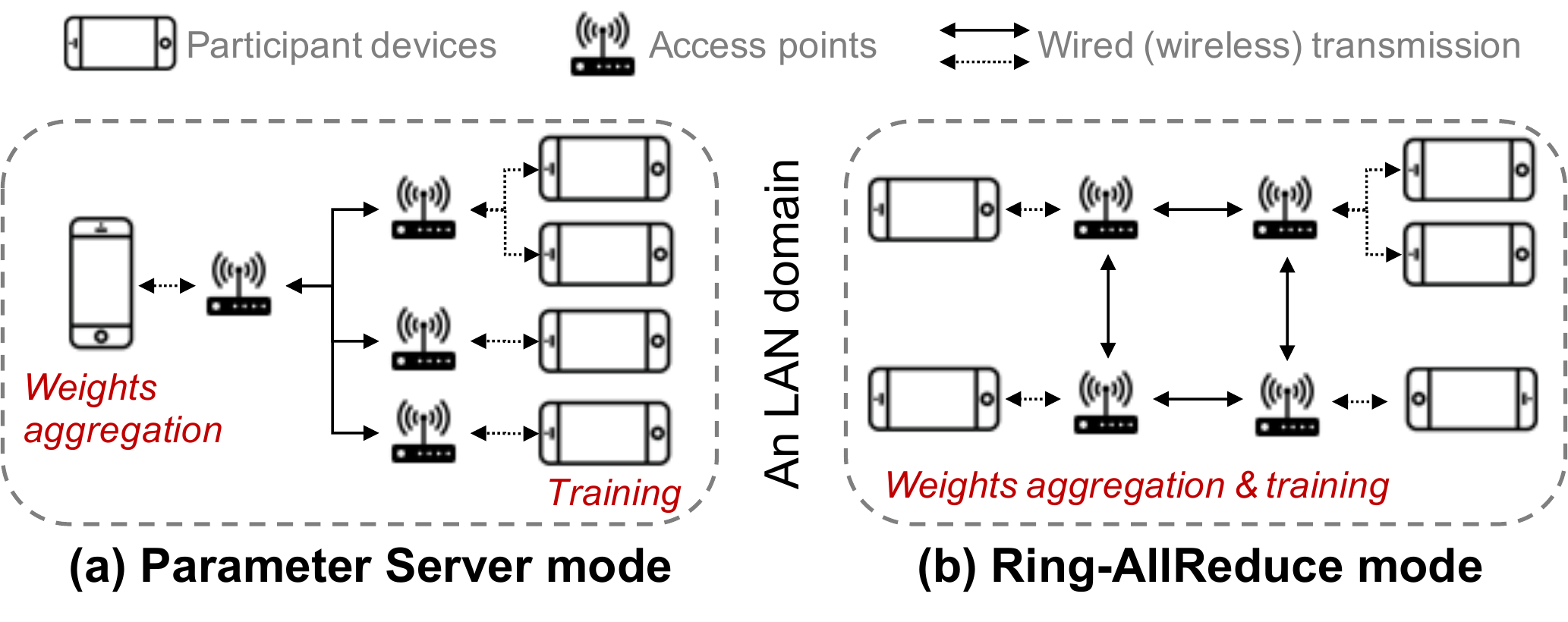}}
\caption{A comparison of two network topologies of how devices within an LAN domain are organized: Parameter Server (PS) mode and Ring-AllReduce mode.}
\label{fig:lan-ps-ring}
\end{figure}

As previously discussed in $\S$\ref{sec:bkgnd}, the network topology about how devices within a LAN domain are organzied has substantial impacts on the network throughput.
For instance, when there is only a few APs, such as in family LAN, the PS mode can obtain higher P2P communication throughput.
When there are more APs and devices in LAN, such as campus LAN, the Ring mode allows for higher throughput.
Figure~\ref{fig:lan-ps-ring} illustrates the comparison of PS and Ring modes.

To obtain a proper topology, the cloud first profiles the network capacity of each AP that are connected to the participant devices at the beginning of FL process.
The capacity information will be used to estimate the P2P transmission throughput under given topology.
When a new cloud round starts, \sys{} first randomly picks the participant devices to guarantee the fairness.
Based on the selected devices, \sys{} iterates over all possible topologies as supported, e.g., PS and Ring for now.
For each topology, \sys{} estimates the P2P throughput for each link, based on the AP capacity and the number of links that are sharing the AP bandwidth simultaneously.
\sys{} then converts the throughput to communication time that such a topology will cost during local aggregation (see below).
Finally, \sys{} picks the toplogy with smallest communication time and employs it for future learning.

We follow the standard way~\cite{patarasuk2009bandwidth, sergeev2018horovod, baidu-allreduce} to simulate the practical collective communication time in each LAN domain under PS and Ring modes.
We denote the communication time in each LAN domain as $com\_T_L$. The $|\omega|$ is the size of global model $\omega$.
For PS mode:
\begin{equation}
\label{eq:ps-design}
   \setlength{\abovedisplayskip}{3pt}
   \setlength{\belowdisplayskip}{3pt}
   com\_T_L = 2 \cdot \frac{|\omega|}{BL_{ps}},
\end{equation}
where $BL_{ps}$ denotes the P2P communication throughput when adopting PS mode.
For Ring mode:
\begin{equation}
\label{eq:ring-design}
   \setlength{\abovedisplayskip}{3pt}
   \setlength{\belowdisplayskip}{3pt}
   com\_T_L = \frac{4(NC_{s}-1)}{NC_s} \cdot \frac{|\omega|}{BL_{ring}},
\end{equation}
where $BL_{ring}$ denotes the P2P communication throughput when adopting Ring mode, and $NC_s$ denotes the number of selected training devices at each device round.
In addition, our $com\_T_L$ in Eq. \ref{eq:ring-design} is 2$\times$ than that in \cite{baidu-allreduce} because the wireless LAN in our \sys{} is half duplex.

\subsection{Dynamic Devices Selection}
\label{subsec:design-heter}

Besides intra-LAN aggregation, \sys{} also relies on cross-LAN coordination to aggregate the updates through WAN.
However, different LAN domains can exhibit disparate bandwidth capacity, called LAN heterogeneity as discussed in $\S$\ref{sec:bkgnd}.
Such heterogeneity leads to out-of-sync training pace of different LANs.
Since \sys{} uses the same training parameters (e.g., device round $RL$) for each LAN domain, the central server has to wait for slow LAN domains to synchronize the generation of new global model.
It inevitably slows down the model convergence.

\sys{} uses the synchronous stochastic gradient descent to update model in parallel, which adopts the bulk synchronous parallel communication mechanism (BSP) \cite{zinkevich2010parallelized}. It is the most widely used algorithm in current FL due to its better accuracy assurance. Therefore, the heterogeneous LAN throughput introduces the straggler problem in \sys{}, which is widely studied in previous works about heterogeneous distributed machine learning \cite{li2014scaling,zinkevich2010parallelized}.  The straggler problem causes the overall performance of \sys{} to be dragged down by the slowest LAN domain, thereby reducing overall acceleration performance.

To fully utilize the computational resources of devices and bandwidth resources within heterogeneous LANs, \sys{} dynamically determines the number of participant devcies for each LAN domain based on its network capacity.
The rationale is that, as previously shown in Figure~\ref{fig:campus-lan}, there exists a trade-off among the device number for parallel training and the average throughput to exchange data between those devices.
Therefore, \sys{} selects more devices to participate in local aggregation for high-bandwidth LAN domains, while less for those slow LAN domains.
In such a way, \sys{} allows heterogeneous LAN domains to perform local aggregation at a consistent pace. 
$\S$\ref{Heterogeneous LAN} will experimentally show the effectiveness of this design.



\section{Evaluation}
\label{sec:eval}

\subsection{Experiment Settings}
\label{subsec:settings}

\begin{table}[t]
\small
\begin{tabular}{lllll}

\textbf{Dataset}   & 
\begin{tabular}[c]{@{}l@{}}\textbf{Model}\\ \textbf{(Size)}\end{tabular} &
\textbf{Task}   & 
\begin{tabular}[c]{@{}l@{}}\textbf{Client}\\ \textbf{number}\end{tabular} & 
\begin{tabular}[c]{@{}l@{}}\textbf{Data per} \\ \textbf{client}\end{tabular} \\ \hline
FEMNIST~\cite{cohen2017emnist} & CNN (25MB)                                             & \begin{tabular}[c]{@{}l@{}}Image clas-\\ sification\end{tabular} & 3550                                                    & 226                                                        \\ \hline
CelebA~\cite{liu2015deep}  & CNN (36MB)                                             & \begin{tabular}[c]{@{}l@{}}Image clas-\\ sification\end{tabular} & 9434                                                    & 21.4                                                       \\ \hline
\end{tabular}
\caption{Datasets and models used in experiments.}
\label{tab:dataset}
\vspace{-0.5cm}
\end{table}

\textbf{Datasets and models} used in our experiments are summarized in Table~\ref{tab:dataset}.
We test \sys{} on 2 datasets: FEMNIST~\cite{cohen2017emnist}, CelebA~\cite{liu2015deep}. We re-use the scripts of LEAF~\cite{leaf} to split each dataset into Non-IID setting and assign them to many devices.
For each dataset, we group the devices into different LAN domains (20 for each by default).
We apply \sys{} on 2 models:
For CelebA, we use a CNN model with 4 CONV, 2 FC, and 84$\times$84 input size.
For FEMNIST, we use a CNN model with 2 CONV, 2 FC, and 28$\times$28 input size.

\noindent
\textbf{Simulation platform and environment}.
We implement \sys{} atop LEAF~\cite{leaf} (TensorFlow-1.13) with two major extensions.
The first one is a communication time module, which simulates the communication time of transmitting models over the WAN and LAN.
The second one is a LAN domain module, which is used to manage the operation of device in each LAN domain, including grouping devices, selecting training devices and the aggregating devices in each LAN (details in $\S$\ref{sec:design}).
All our experiments are performed on a high-performance server with Ubuntu 18.04. The server has 8$\times$ NVIDIA RTX 2080Ti GPUs with sufficient GPU memory.

\noindent
\textbf{FL configurations}. We mainly follow prior FL literature~\cite{leaf, yang2020heterogeneity, wang2020federated} to set the default configurations in our experiments, such as learning rate and batch size.
We fix the $BW$=2Mbps and $N_s$=50 (the number of selected training devices) in WAN-FL.
To evaluate the impact of $BL$, $NL_s$ (the number of selected LAN domain) and $NC_s$ (the number of selected training devices in each LAN domain) on \sys{}, we try different values of $BL$ as 5-40Mbps, $NL_s$ as 1-10 and $NC_s$ as 2-20 respectively.
We will provide the detailed settings in following subsections.

\noindent
\textbf{Baseline}
In all experiments, we treat the original LEAF~\cite{leaf} as the baseline, which is termed as WAN-FL.
Its training parameters are set to be consistent with \sys.
\subsection{Evaluation Metrics}
\label{sec:eval-metrics}

\textbf{Accuracy} We divide the dataset of each device into training set (80\%) and testing set (20\%). We test the finally global model on the combined testing set from all devices.
A model convergence is defined when the variance of accuracy is less than 0.2\% for 20 consecutive cloud rounds \cite{goyal2017accurate}.

\noindent
\textbf{Clock time} We characterize the $clock\_ time$ as the total time spent during FL, including the on-device training time ($train\_T_C$) and the communication time across WAN and LAN, respectively denoted by $com\_T_W$ and $com\_T_L$.
To obtain $train\_T_C$, we follow prior work~\cite{yang2020heterogeneity} to profile the training performance of each model using DL4J~\cite{dl4j}, an open-source ML library supporting Android devices.
Based on the \sys{} design, the clock time can be obtained as:
\begin{equation}
   \setlength{\abovedisplayskip}{3pt}
   \setlength{\belowdisplayskip}{3pt}
   clock\_time = \sum_{RW}(com\_T_W + \sum_{RL}(train\_T_C + com\_T_L)).
\end{equation}
The $com\_T_W$ can be estimated by the size of global model ($|\omega|$) and the network bandwidth ($BW$).
The $com\_T_L$ can be calculated as the Eq. \ref{eq:ps-design} or Eq. \ref{eq:ring-design} in $\S$\ref{subsec:lan-aware-cc}.

\noindent
\textbf{WAN traffic} We quantify the $WAN\_traffic$ in terms of total traffic from cloud to devices during the FL process till model convergence.
For \sys{}, it can be calculated as:
\begin{equation}
   \setlength{\abovedisplayskip}{3pt}
   \setlength{\belowdisplayskip}{3pt}
   WAN\_traffic = RW \cdot NL_s \cdot |\omega|,
\end{equation}
where $NL_s$ denotes the number of selected LAN domains participating in \sys{}.
For WAN-FL, it can be calculated as:
\begin{equation}
   \setlength{\abovedisplayskip}{3pt}
   \setlength{\belowdisplayskip}{3pt}
   WAN\_traffic = RW \cdot N_s \cdot |\omega|,
\end{equation}
where $N_s$ denotes the number of selected training devices in each cloud round $RW$.

\noindent
\textbf{Monetary cost} is estimated as if deploying \sys{} on AWS EC2.
We use a1.2xlarge instance~\cite{aws-price} with 8$\times$vCPU and 16GB memory, which is enough for model aggregation and costs \$0.204/hour.
The uplink traffic is free and the downlink cost is \$0.09/GB.
Thus, The cost is calculated as:
\begin{equation}
\label{eq:money}
   \setlength{\abovedisplayskip}{3pt}
   \setlength{\belowdisplayskip}{3pt}
   Cost = 0.204 \cdot clock\_time + 0.09 \cdot WAN\_traffic.
\end{equation}
\subsection{End-to-end Results}
\label{End-to-end Results}


\begin{table*}[t]
\centering
\small
\begin{tabular}{|l|l|l|l|l|l|l|}
\hline
Dataset                   & FL Protocols                                                 & Accuracy (\%)                        & Round Number                         & WAN Traffic (GB)                     & Clock Time (hour)                    & Monetary Cost (\$)                         \\ \hline
                          & WAN-FL (E=10)                                      & 81.82                                                   & 1820                                         & 2221                                          & 170                               & 234.57                                 \\ \cline{2-7}
                          & WAN-FL (E=1)                                       & 82.92 (1.10$\uparrow$)                          & 5000 (274\%)                            & 6103 (274\%)                             & 298 (175\%)                                & 610.06 (260\%)                                 \\ \cline{2-7}
                          & WAN-FL (E=50)                                     & 81.43 (0.39$\downarrow$)                    & 1660 (91\%)                              & 2026 (91\%)                               & 342 (201\%)                                & 252.11 (107\%)                                \\ \cline{2-7}
\multirow{-3}{*}{FEMNIST} & \cellcolor[HTML]{DAE8FC}\textbf{LanFL (E=2, RL=5)}  & \cellcolor[HTML]{DAE8FC}\textbf{82.85 (1.03$\uparrow$)} & \cellcolor[HTML]{DAE8FC}\textbf{240 (13\%)} & \cellcolor[HTML]{DAE8FC}\textbf{29 (1\%)} & \cellcolor[HTML]{DAE8FC}\textbf{28 (16\%)} & \cellcolor[HTML]{DAE8FC}\textbf{8.32 (3\%)}  \\ \hline
                          & WAN-FL (E=20)                                      & 91.56                                  & 800                         & 1406                       & 208                        & 169.24                                 \\ \cline{2-7}
                          & WAN-FL (E=1)                                       & 91.77 (0.21$\uparrow$)       & 1600 (200\%)          & 2812 (200\%)           & 143 (69\%)            & 282.25 (167\%)                                 \\ \cline{2-7}
                          & WAN-FL (E=50)                                      & 91.43 (0.13$\downarrow$)   & 760 (95\%)             & 1335 (95\%)              & 403 (194\%)         & 150.61 (89\%)                                \\ \cline{2-7}
\multirow{-3}{*}{CelebA}  & \cellcolor[HTML]{DAE8FC}\textbf{LanFL (E=2, RL=10)} & \cellcolor[HTML]{DAE8FC}\textbf{91.66 (0.10$\uparrow$)} & \cellcolor[HTML]{DAE8FC}\textbf{420 (53\%)} & \cellcolor[HTML]{DAE8FC}\textbf{73 (5\%)} & \cellcolor[HTML]{DAE8FC}\textbf{141 (68\%)} & \cellcolor[HTML]{DAE8FC}\textbf{35.33 (21\%)} \\ \hline
\end{tabular}
\caption{Summarized performance of \sys{} compared to WAN-FL.}
\label{tab:result-time}
\end{table*} 


\begin{figure}[t]
	\centering
	\begin{minipage}[b]{0.24\textwidth}
	\includegraphics[width=1.0\textwidth]{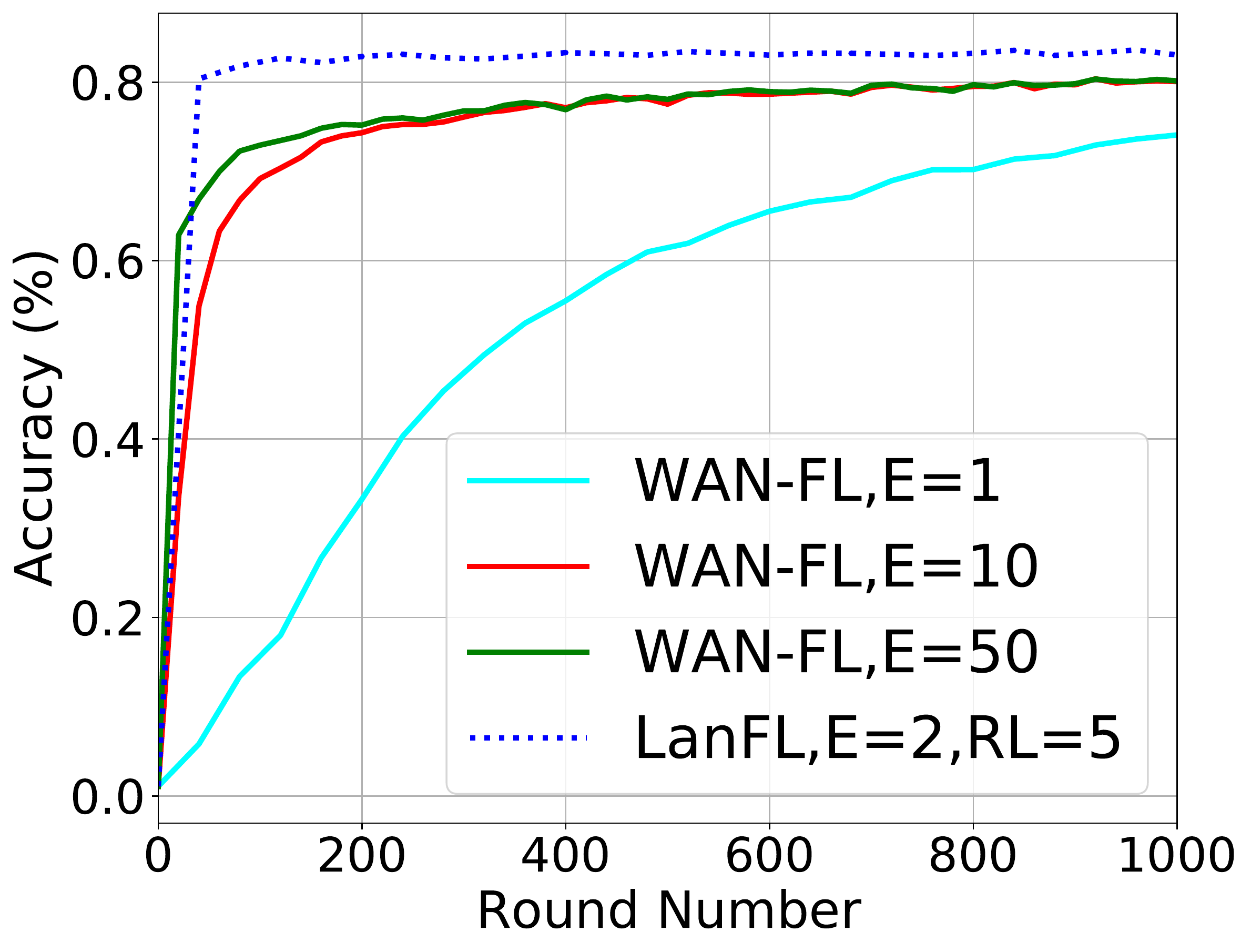}
	\subcaption{Acc. vs. round (FEMNIST)}
	\end{minipage}
    ~
    \begin{minipage}[b]{0.24\textwidth}
	\includegraphics[width=1.0\textwidth]{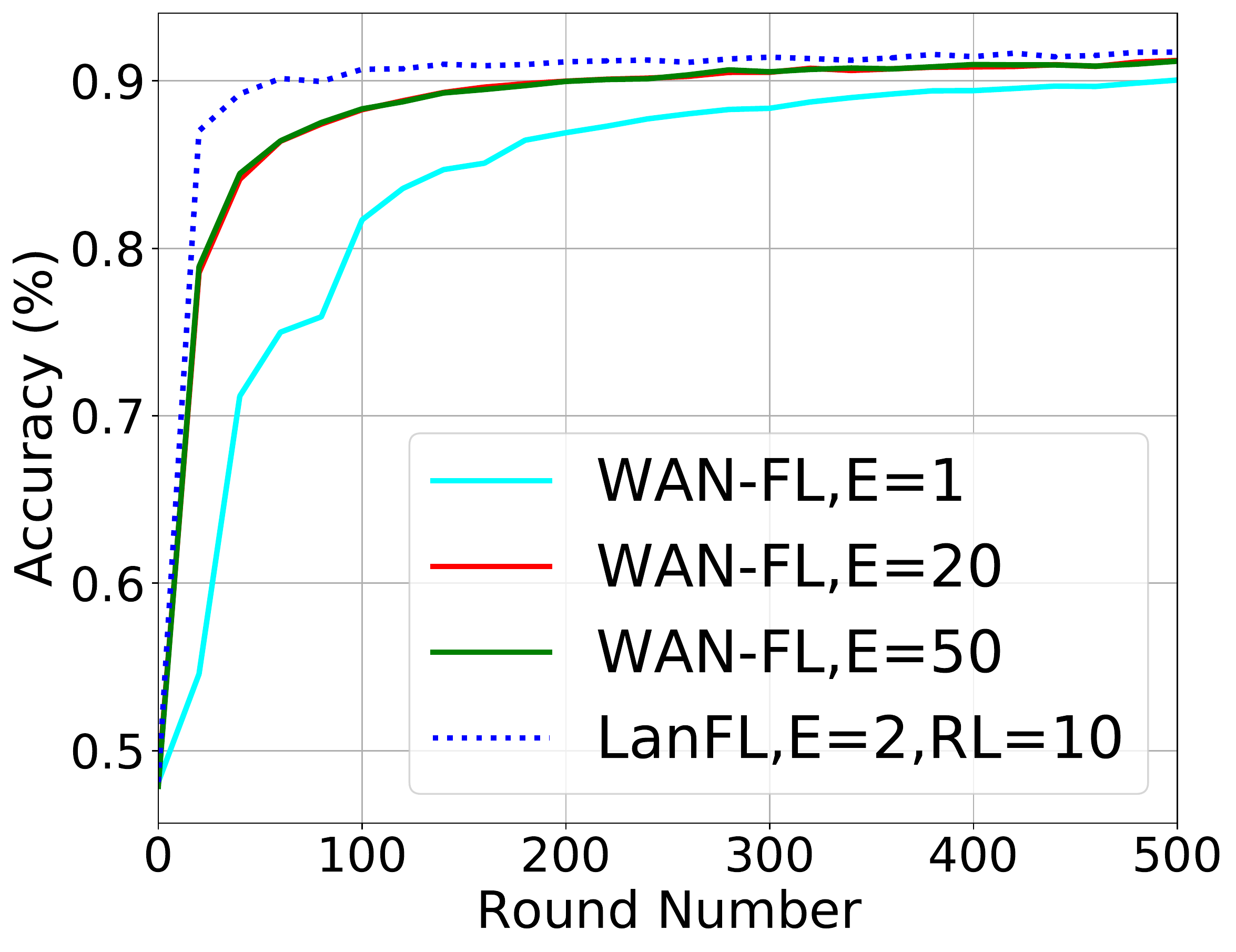}
	\subcaption{Acc. vs. round (CelebA)}
	\end{minipage}

	\begin{minipage}[b]{0.24\textwidth}
	\includegraphics[width=1.0\textwidth]{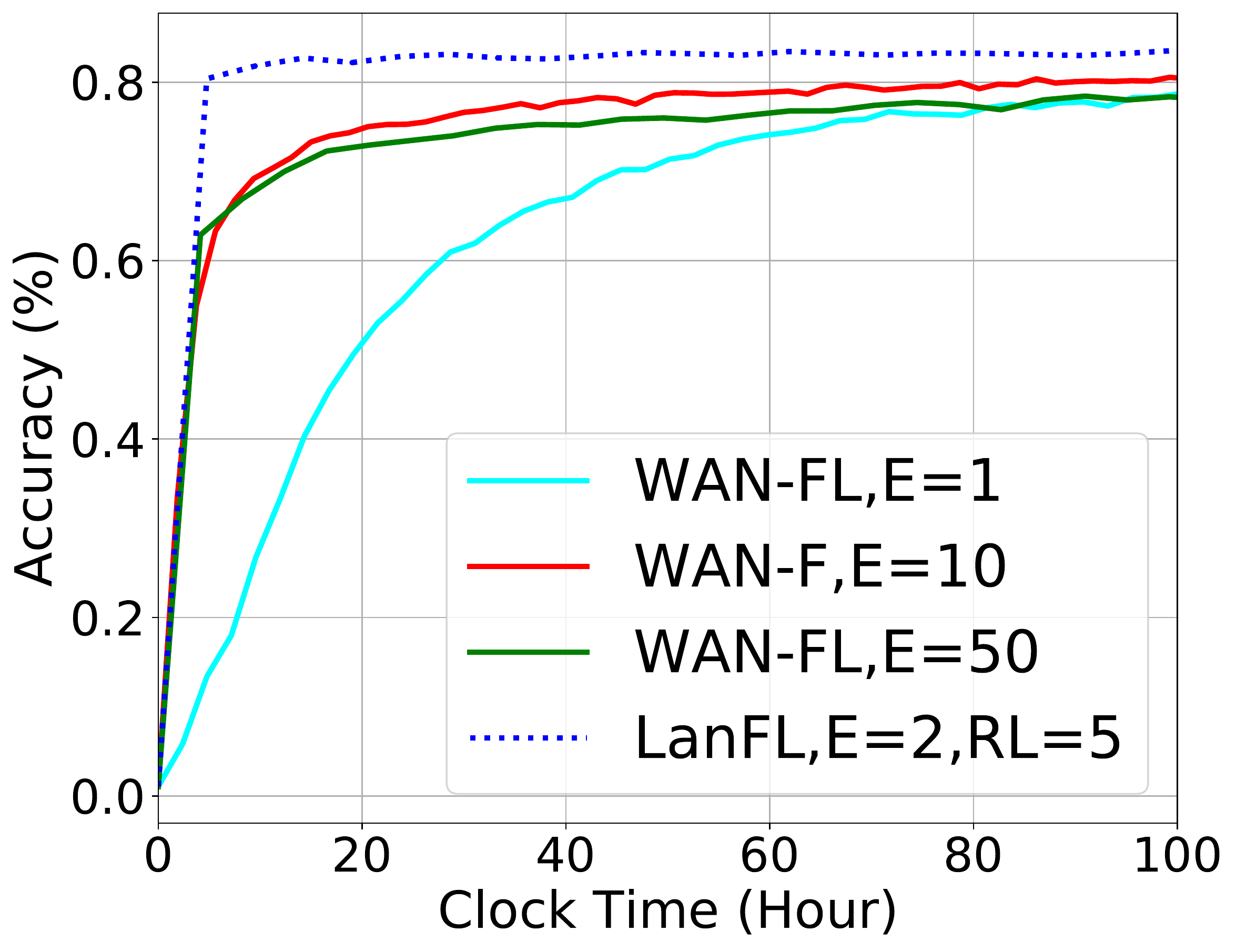}
	\subcaption{Acc. vs. time (FEMNIST)}
	\end{minipage}
    ~
    \begin{minipage}[b]{0.24\textwidth}
	\includegraphics[width=1.0\textwidth]{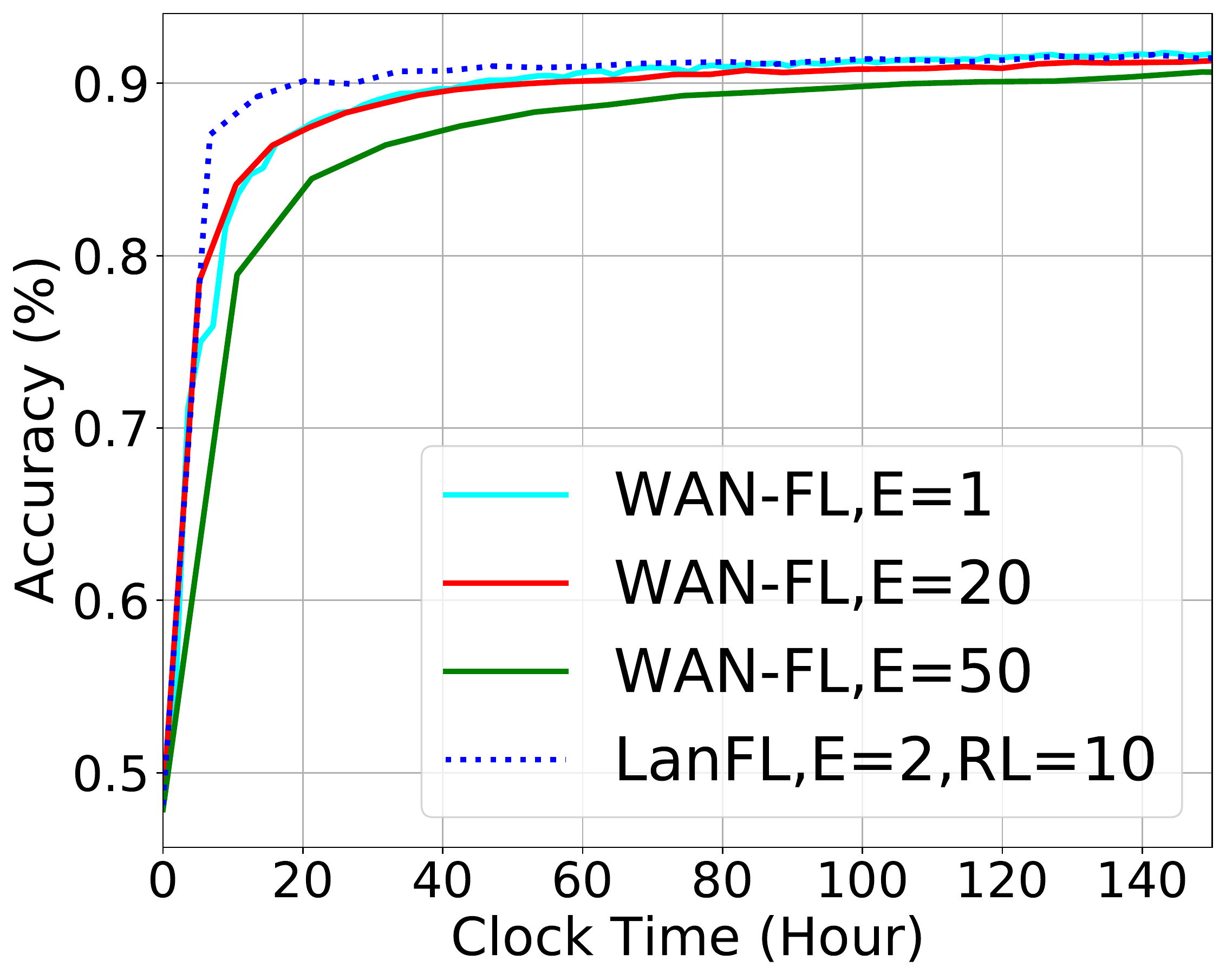}
	\subcaption{Acc. vs. time (CelebA)}
	\end{minipage}
    \caption{The testing accuracy across (Acc.) training rounds and clock time on two datasets.}
    \label{fig:acc-time-round-leaf}
\end{figure} 

We first present the overall results of \sys{} compared with WAN-FL.
Table~\ref{tab:result-time} summarizes the key metrics when model converges and Figure~\ref{fig:acc-time-round-leaf} illustrates the convergence process.
For WAN-FL, we vary the local epoch number ($E$) for its key role in tuning FL performance ($\S$\ref{subsec:LAN-design}).
For each device round in LAN-FL, \sys{} selects 5 LAN domains ($NL_s$), and each domain involves 10 devices for training ($NC_s$).
The WAN and LAN domain bandwidths are set to 2Mbps and 20Mbps, respectively.
Noting that this LAN bandwidth is the final network throughput after being shared across devices instead of the AP bandwidth capacity.
We will further evaluate \sys{} with more settings in following subsections.

\textbf{\sys{} can significantly accelerate the model convergence while preserving model accuracy in FL}.
Treating WAN-FL ($E$=10 for FEMNIST, $E$=20 for CelebA) as baseline,
reducing $E$ to 1 increases the accuracy (1.10\% for FEMNIST, 0.21\% for CelebA) but takes much longer clock time towards convergence (1.75$\times$ for FEMNIST, 1.1$\times$ for CelebA).
Using a larger epoch size ($E$=50) cannot improve the learning process as well.
However, \sys{} improves both the accuracy and clock time compared with baseline, especially the clock time.
On FEMNIST,  the accuracy of \sys{} ($E$=2, $RL$=5) is 1.03\% higher than baseline, and it only takes 16\% of clock time (i.e., 6.25$\times$ speedup).
On CelebA, the accuracy of \sys{} ($E$=2, $RL$=10) is only 0.10\% higher than baseline, but it takes 68\% of clock time (i.e., reducing 67 hours).

\textbf{\sys{} can significantly reduce WAN traffic and monetary cost}.
Figure \ref{fig:acc-time-round-leaf} shows that the number of cloud round $RW$ required by \sys{} to converge is much fewer than baseline (87\% on FEMNIST, 47\% on CelebA), which brings in tremendous WAN traffic savings.
Table \ref{tab:result-time} also shows that the WAN traffic and monetary cost of \sys{} is much less than baseline:
On FEMNIST, \sys{} reduces 99 \% WAN traffic and 97\% monetary cost to the baseline;
On CelebA,  \sys{} reduces 95\% WAN traffic and 79\% monetary cost to the baseline.
The saved cost comes from both the reduced clock time to rent hardware resources and the reduced network traffic according to Eq. \ref{eq:money}.

The tremendous improvement of \sys{} comes from its LAN-aware design.
\sys{} introduces a LAN domain that uses a small local epochs $E$ and a large device round $RL$. On one hand, this design can reduce the bias between LAN domains in each cloud round $RW$ compared with using a large $E$ in WAN-FL. On the other hand, it can reduce the cloud round $RW$ across WAN compared with using a small $E$ in WAN-FL.
Moreover, the number of devices that need to exchange model weights with the central cloud in WAN-FL is much more than \sys{} explained in $\S$~\ref{subsec:workflow}.


\begin{figure}[t]
	\centering
	\begin{minipage}[b]{0.24\textwidth}
		\includegraphics[width=1.0\textwidth]{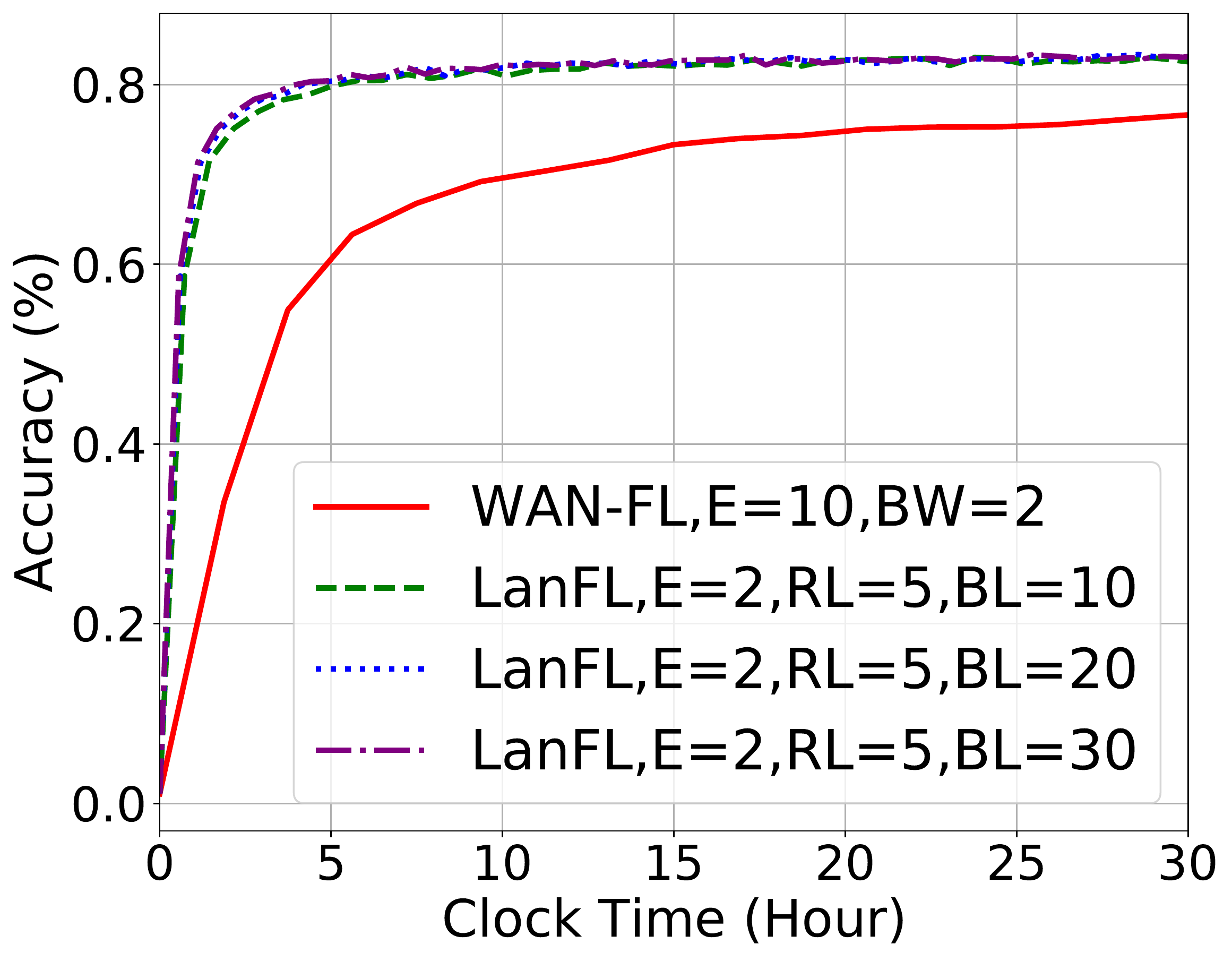}
		\subcaption{FEMNIST}
	\end{minipage}
	~
	\begin{minipage}[b]{0.24\textwidth}
	\includegraphics[width=1.0\textwidth]{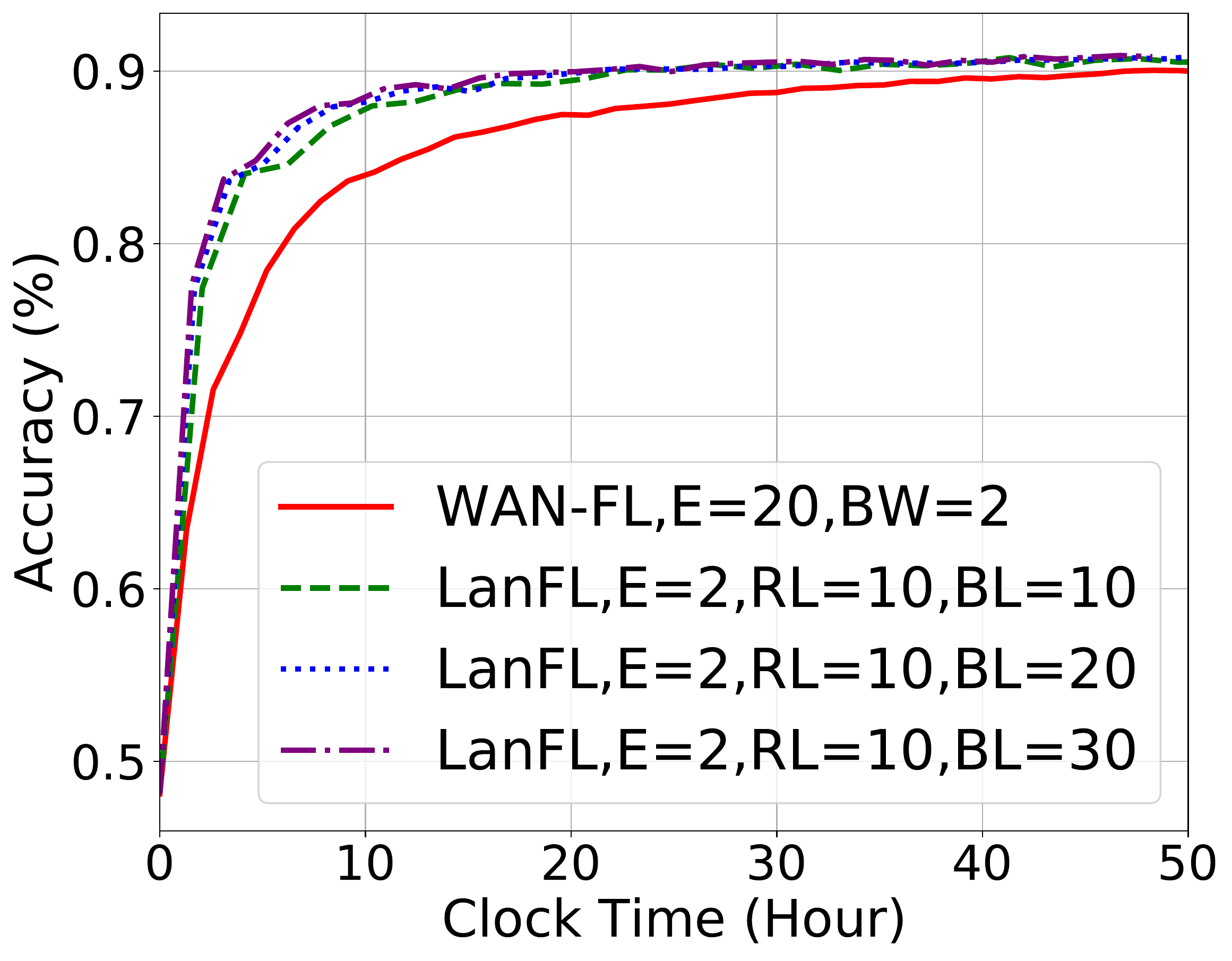}
	\subcaption{CelebA}
	\end{minipage}
    \caption{The training performance of \sys{} and WAN-FL across different LAN network throughputs.}
    \label{fig:eval-time-speed-lan}
\end{figure}

\textbf{Impact of LAN bandwidth}.
We further investigate the performance of \sys{} under various LAN bandwidth $BL$ (10Mbps, 20Mbps, and 30Mbps).
We choose the WAN-FL ($E$=20 for CelebA and $E$=10 for FEMNIST) as our baseline due to their better trade-off on weighted accuracy and clock time.
As illustrated in Figure \ref{fig:eval-time-speed-lan} that the training speed can be further improved by increasing the throughput of P2P communication links in LAN domain.
For example, it takes \sys{} 3.85 hours to reach 80\% accuracy on FEMNIST with $BL$=30Mbps, which is 1.2$\times$ faster than \sys{} with $BL$=10Mbps.
On CelebA, it takes \sys{} 21.77 hours to reach 90\% accuracy with $BL$=30Mbps, which is 1.3$\times$ faster than \sys{} with $BL$=10Mbps.
We notice that the training speed cannot proportionably boost with LAN bandwidth improvement, because on-device training also takes substantial time.
\subsection{Analysis of LAN and Device Number}
\label{eval:number}

In this section, we study the performance of \sys{} with different number of selected LAN domains ($NL_s$=1,2,5,10) and the number of selected training devices in each LAN domain ($NC_s$=2,5,10,20).
Other settings of \sys{} are consistent with experiments in Figure \ref{fig:acc-time-round-leaf}.


\begin{figure}[t]
	\centering
    \begin{minipage}[b]{0.24\textwidth}
	\includegraphics[width=1.0\textwidth]{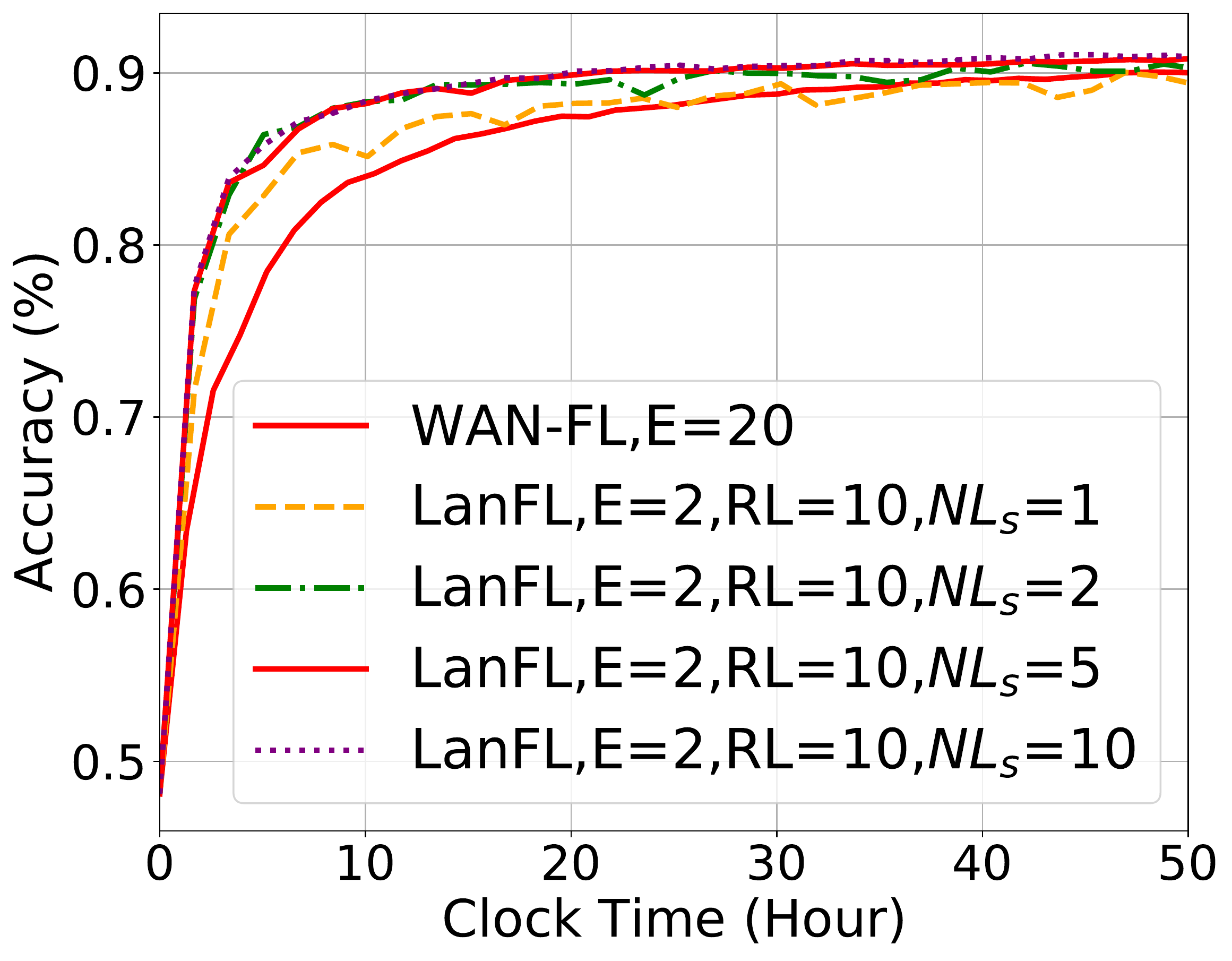}
	\subcaption{Various domain number}
	\end{minipage}
    ~
    \begin{minipage}[b]{0.24\textwidth}
	\includegraphics[width=1.0\textwidth]{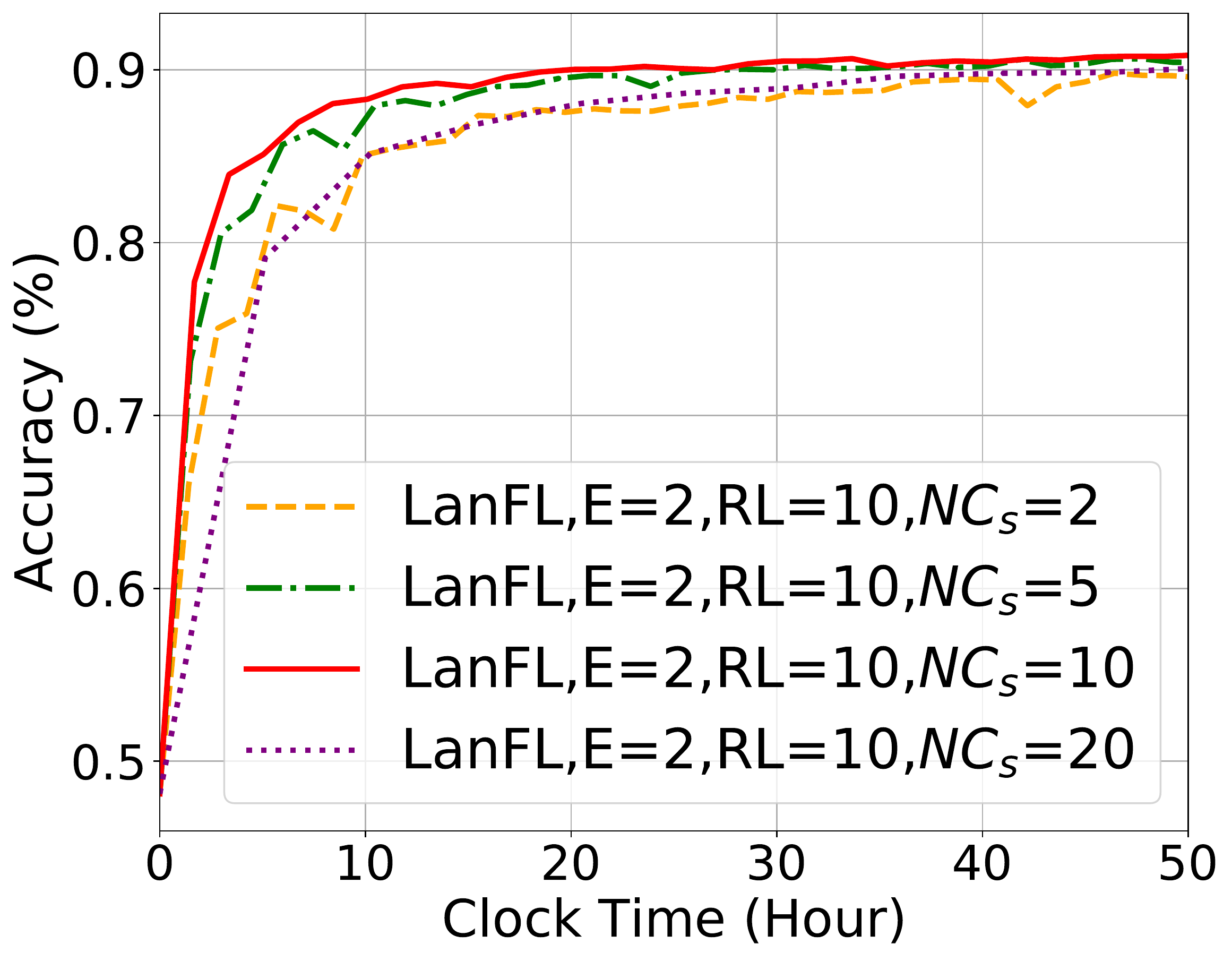}
	\subcaption{Various device number}
	\end{minipage}
    \caption{\sys{}'s  performance across different (a) \# of LAN domains and (b) \# of devices per domain. Both experiments are performed on CelebA dataset.}
    \label{fig:eval-time-num-lan}
\end{figure}

\textbf{LAN domain number}
As shown in Figure \ref{fig:eval-time-num-lan} (a), using only 1 LAN domain results in lower convergence accuracy, e.g., 1.53\% drop compared to 5 LAN domains.
It confirms the hierarchical design of \sys{} involving both intra-LAN and inter-LAN collaborative training.
Increasing the LAN domain number can boost the training, e.g., 1.5$\times$ faster towards 90\% convergence accuracy for 5 domains against 2 domains.
However, further increasing the LAN domain number, e.g., from 5 to 10, can barely accelerate the training speed.
Such an observation is consistent with original FL protocol~\cite{bonawitz2019towards} where the training performance cannot be improved once the participant device number reaches a throttle.

\textbf{Per-domain device number}
Figure \ref{fig:eval-time-num-lan} (b) shows that, with 10 devices participating in local aggregation per LAN domain ($NC_s$=10), \sys{} obtains the optimal training performance.
For example, to reach 90\% convergence accuracy, it is 2.3$\times$ and 2.2$\times$ faster than $NC_s$=2 and $NC_s$=20, respectively.
This is because, when training device number is small (i.e., 2), the invovled training data per local round is small and the aggregated model is likely to be biased.
While the training device number is too large (i.e., 20), the P2P communication throughput within LAN domains are decreased and thus the local aggregation is slowed down.
It shows an interesting yet unexplored trade-off opened by LAN sharing among collaborative training devices as discussed in $\S$\ref{sec:bkgnd}.



\subsection{Analysis of LAN topology}
\label{subsec:eval-topology}


\begin{figure}[t]
	\centering
	\begin{minipage}[b]{0.24\textwidth}
	\includegraphics[width=1.0\textwidth]{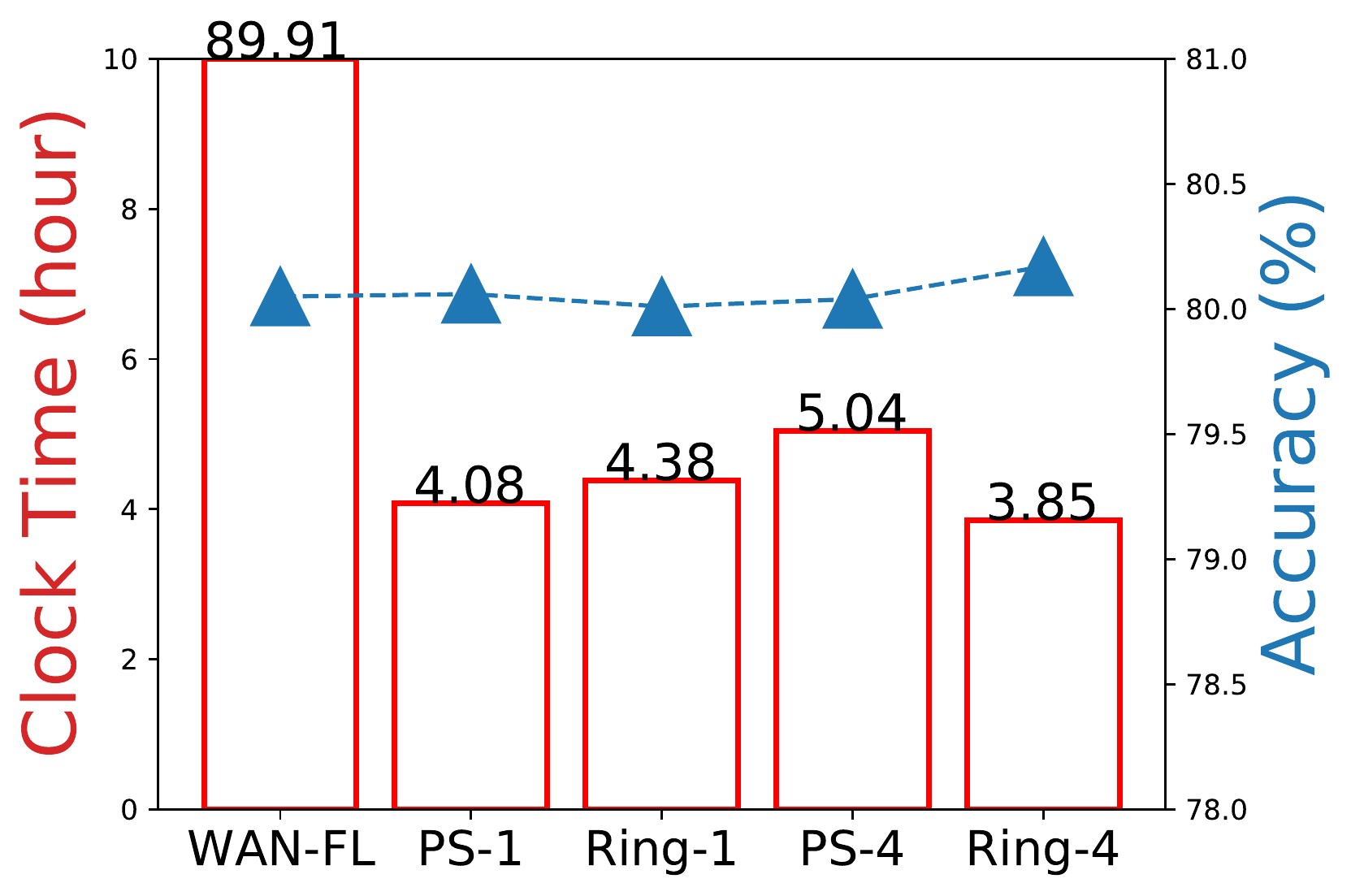}
	\subcaption{FEMNIST}
	\end{minipage}
    ~
    \begin{minipage}[b]{0.24\textwidth}
	\includegraphics[width=1.0\textwidth]{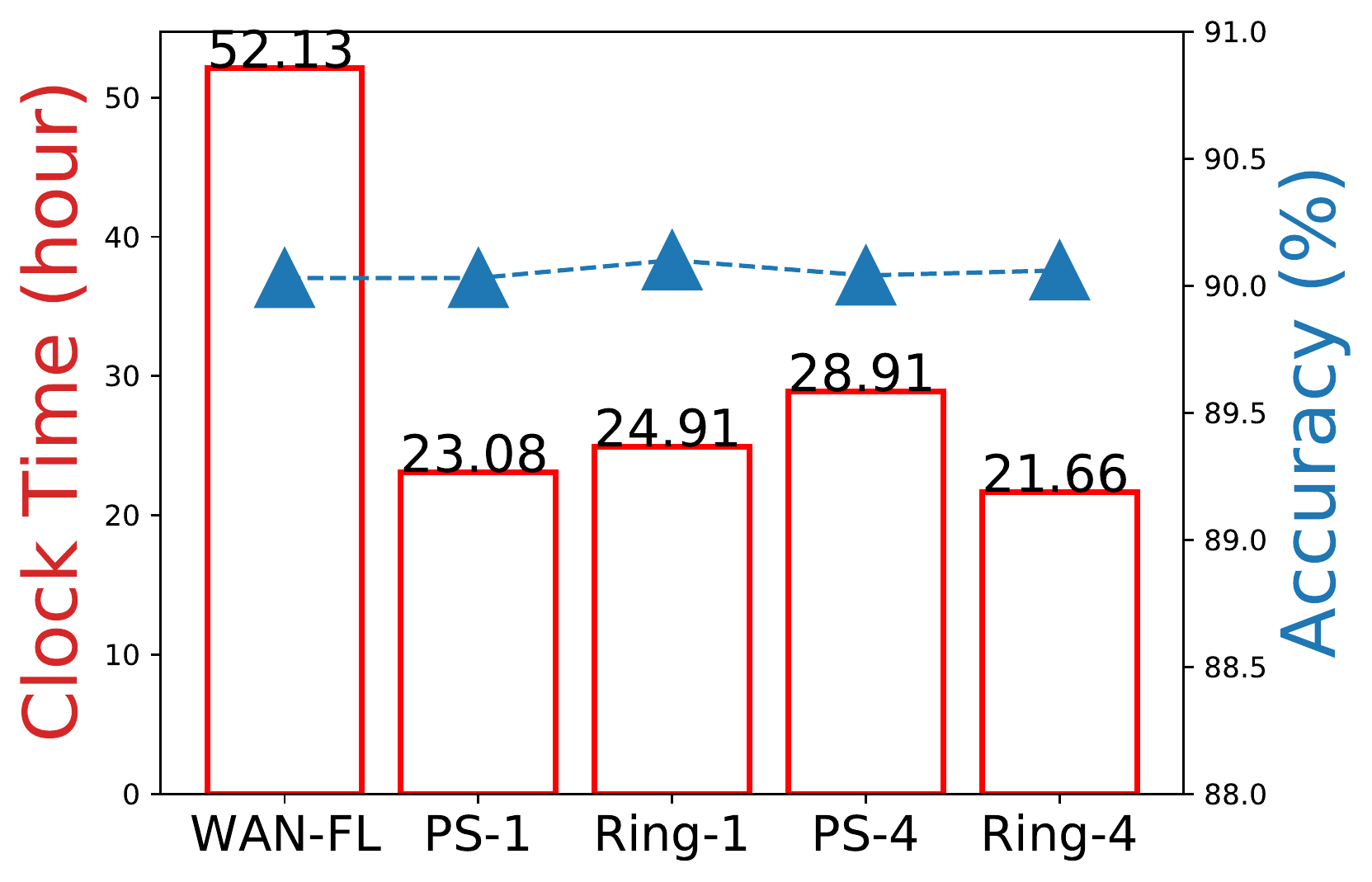}
	\subcaption{CelebA}
	\end{minipage}
    \caption{The training performance of WAN-FL and \sys{} with different PS and Ring mode.}
    \label{fig:eval-time-ps-ring}
\end{figure} 

We further evaluate \sys{}'s design in determining intra-LAN device topology ($\S$\ref{subsec:lan-aware-cc}) by testing \sys{} with both PS and Ring modes.
We choose WAN-FL with the same settings as Figure \ref{fig:eval-time-speed-lan}.
We set the configurations of \sys{} as in Figure \ref{fig:eval-time-ps-ring} according to our measured results shown in Figure \ref{fig:campus-lan}.
We choose the four measured average throughput of 8 devices in Figure \ref{fig:campus-lan} as the $BL$ of \sys{}.
We denote the four cases in Figure \ref{fig:campus-lan} as PS-1, PS-4, Ring-1 and Ring-4, and their throughput are 22Mbps, 16Mbps, 20Mbps and 62Mbps, respectively.
Other settings are consistent with experiments in Figure \ref{fig:acc-time-round-leaf}.

The results are illustrated in Figure \ref{fig:eval-time-ps-ring}, from which we make following key observations.
First, \sys{} outperforms the baseline with either PS or Ring mode, e.g.,  22.0$\times$--23.4$\times$ and 2.3$\times$--2.4$\times$ faster respectively at accuracy 80\% on FEMNIST and 90\% on CelebA.
Second, \sys{} adopts PS mode which outperforms Ring mode when AP number is small.
For example, on FEMNIST and 1 AP, \sys{} takes 1.2$\times$ longer time till model convergence.
When the AP number is larger, however, \sys{} favors Ring mode, e.g., 1.1$\times$ faster than PS mode with 4 APs on FEMNIST.
It shows that \sys{} can judiciously pick the optimal network topology due to its LAN-aware design.

\subsection{Analysis of Heterogeneous LAN}
\label{Heterogeneous LAN}


\begin{figure}[t]
	\centering
	\begin{minipage}[b]{0.24\textwidth}
	\includegraphics[width=1.0\textwidth]{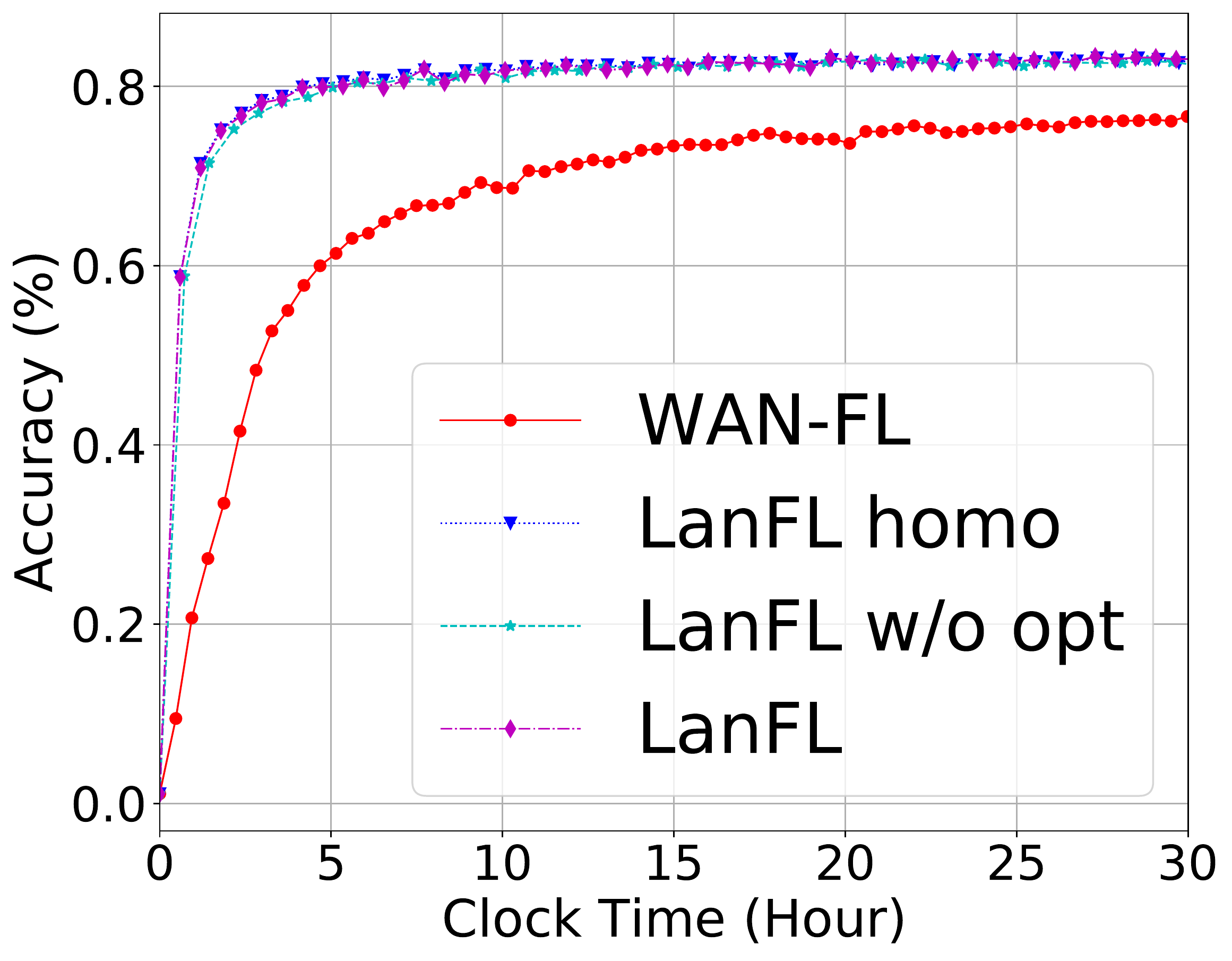}
	\subcaption{FEMNIST - [10,10,10,10,30]}
	\end{minipage}
	~
	\begin{minipage}[b]{0.24\textwidth}
	\includegraphics[width=1.0\textwidth]{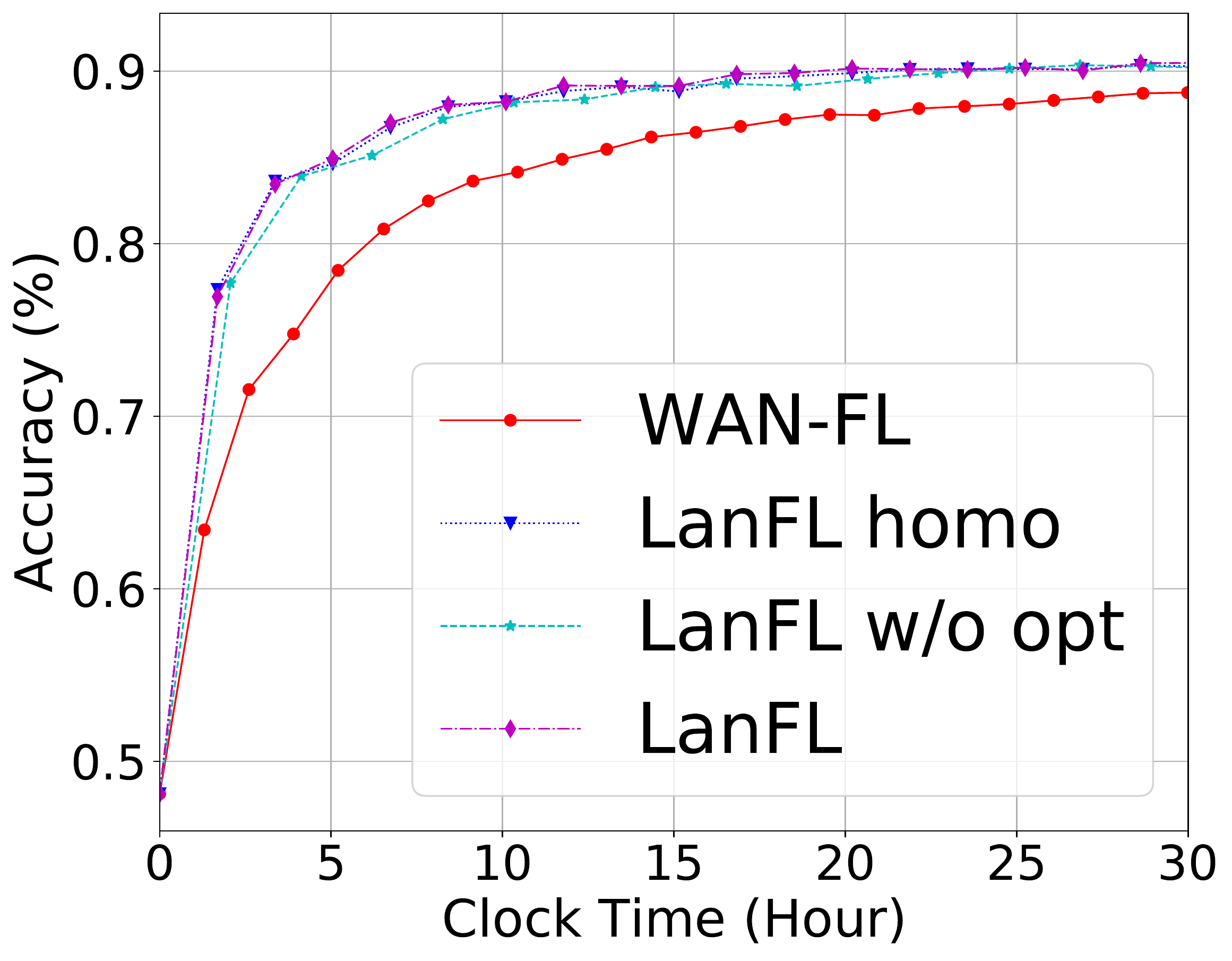}
	\subcaption{CelebA - [10,10,10,10,30]}
	\end{minipage}

    \begin{minipage}[b]{0.24\textwidth}
	\includegraphics[width=1.0\textwidth]{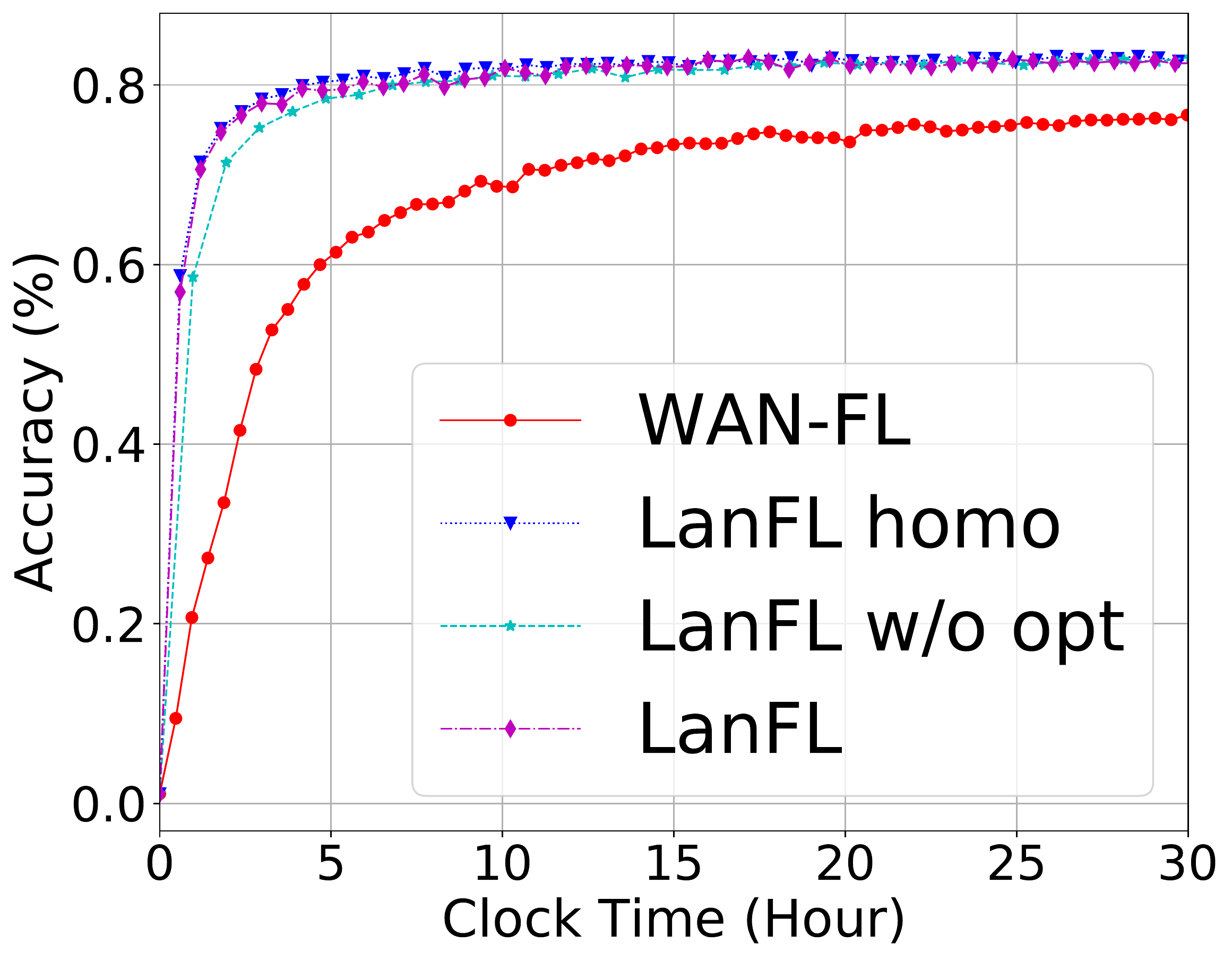}
	\subcaption{FEMNIST - [5,20,20,40,40]}
	\end{minipage}
	~
	\begin{minipage}[b]{0.24\textwidth}
	\includegraphics[width=1.0\textwidth]{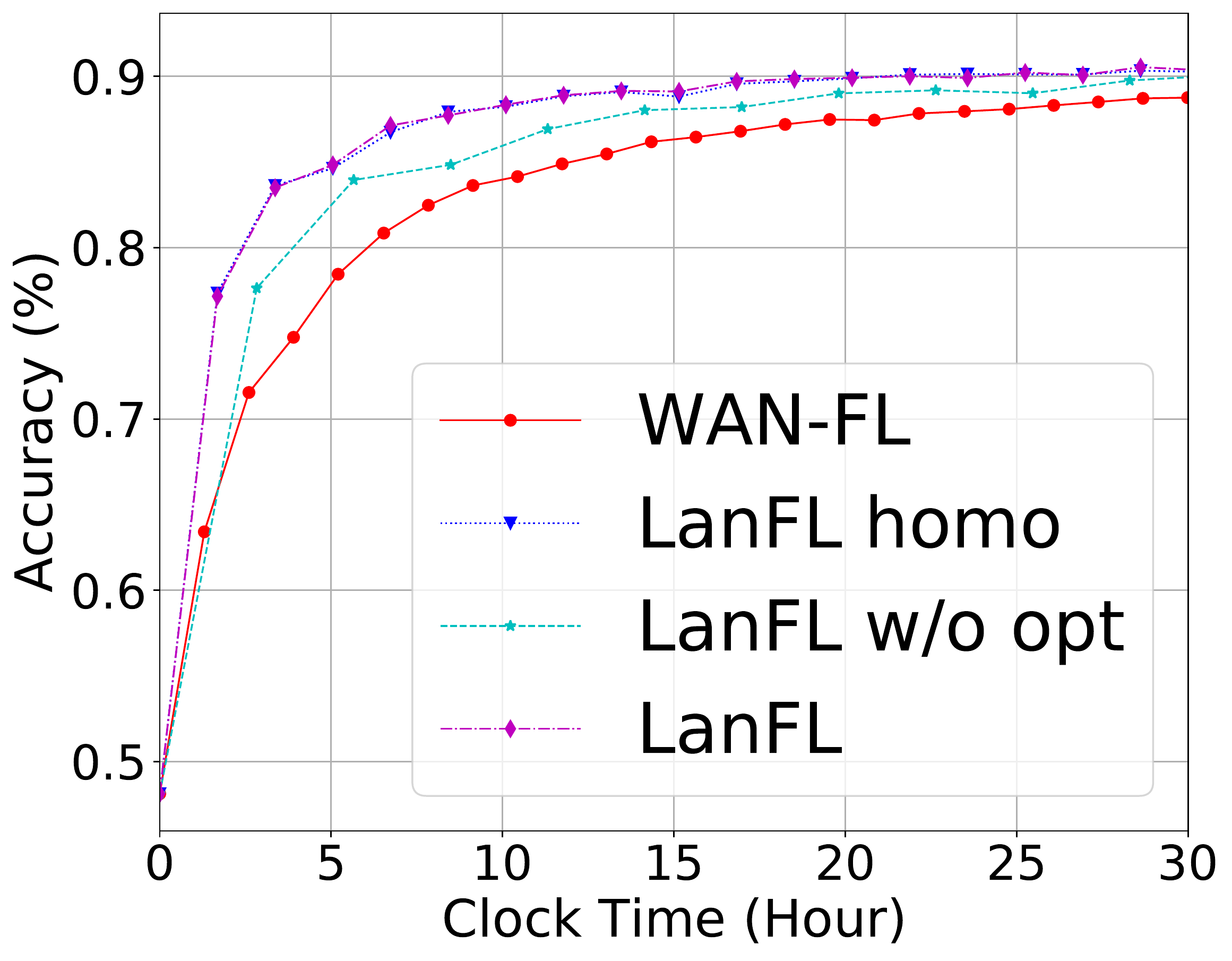}
	\subcaption{CelebA - [5,20,20,40,40]}
	\end{minipage}
	\caption{The training performance of \sys{} and baselines under heterogeneous LAN settings. The list of following numbers are the LAN bandwidths (Mbps).}
	\label{fig:eval-heter}
\end{figure} 

We further evaluate \sys{}'s design in dealing with heterogeneous LAN.
We randomly try different heterogeneous $BL$ settings for 5 LAN domains: (10,10,10,10,30) and (5,20,20,40,40) (unit: Mbps) and the results are shown in Figure \ref{fig:eval-heter}(a, b) and (c, d), respectively.
We use three baselines:
(1) WAN-FL,
(2) \sys{} under homogeneous LANs with the average bandwidth as all LAN domains (denoted as ``\sys{} homo'')
(3) a naive version of \sys{} (denoted as ``\sys{} w/o opt'') that waits and synchronizes for the slow LAN domains without dynamic device selection optimization ($\S$\ref{subsec:design-heter}).
The settings are consistent with previous experiments in Figure \ref{fig:acc-time-round-leaf}.

\textbf{LAN heterogeneity slows down model convergence for LAN-aware FL as compared to homogeneous setting.}
In Figure \ref{fig:eval-heter} (c, d) with (5,20,20,40,40) bandwidth, \sys{} without optimization takes 1.6$\times$ and 1.7$\times$ longer clock time to converge for FEMNIST and CelebA datasets as against homogeneous setting, respectively.
Under the other bandwidth setting, shown in Figure \ref{fig:eval-heter} (a, b), the clock time is also lengthened by 1.2$\times$ and 1.3$\times$, respectively.
The reason of such performance degradation is because, without optimization, \sys{} is bottlenecked by the slowest LAN domain due to the BSP mechanism~\cite{zinkevich2010parallelized} adopted by \sys.
Noting that, even without optimization, \sys{} still outperforms WAN-FL.


\textbf{Through dynamic device selection, \sys{} can effectively mitigate the performance degradation caused by LAN heterogeneity.}
For example, under (5,20,20,40,40) bandwidth setting, \sys{} with optimization can achieve almost the same performance as \sys{} in CelebA.
We zoom into the experiments and find that \sys{} picks 2, 10, 10, 14, 14 devices for each LAN domain, respectively, to balance the training pace among heterogeneous LAN domains.
On FEMNIST, \sys{} is not as good as homogeneous setting, but it still improves the convergence speed by 14.71\% -- 54.6\% as shown in Figure \ref{fig:eval-heter}.
It shows that \sys{}'s design can narrow down the bandwidth gap between heterogeneous LAN domains, and thus reduce the waiting time on the slow LAN domain.
\section{Related Work}
\label{sec:related}

\noindent
\textbf{Federated learning (FL)} is an emerging machine learning paradigm to enable many clients to collaboratively train a ML model while preserving data privacy ~\cite{konevcny2016federatedoptimization, konevcny2016federatedlearning, mcmahan2017communication}.
A typical FL process involves many clients that hold the Non-IID datasets and train the model separately, and one centralized server that aggregates the model updates from clients.
This work targets cross-device FL, the classic FL scenario where each client is a mobile device~\cite{hard2018federated, yang2018applied, chen2019federated, ramaswamy2019federated, leroy2019federated}.
A broader definition of FL~\cite{kairouz2019advances} also includes cross-silo FL where each clients are different organizations (e.g., medical or financial) or geo-distributed datacenters~\cite{courtiol2019deep}. The popularity of federated learning technology has also resulted in a number of tools and frameworks~\cite{ryffel2018generic,leaf,yang2020heterogeneity}.

\noindent \textbf{Edge-assisted hierarchical FL}
is recently proposed~\cite{lim2020federated, liu2019edgeassist, abad2020hierarchical, briggs2020federated} to reduce the communication rounds between remote cloud and clients.
Those preliminary efforts mostly use edge servers to aggregate the model updates from nearby clients.
Compared to our LAN-driven FL proposal, those approaches are impractical given that public edge servers are not commonly available nowadays (e.g., only LA for AWS Local Zones~\cite{aws-local-zones} in U.S.) and, even if possible, it cannot speed up FL convergence because the edge servers are still bandwidth-constrained as clouds.
Furthermore, the usage of many edge servers could incur even more monetary cost to FL developers than traditional FL.


\noindent \textbf{FL communication optimization} is the primary goal of this work.
All existing related literatures are to reduce WAN communication in FL.
Some of them~\cite{li2018federated,smith2017cocoa,stich2018local,zhang2015deep} focus on reducing the total number of communication rounds.
These mini-batch optimization methods are widely used in distributed machine learning at a centralized data center.
In practice, they have shown limited flexibility to adapt to communication-computation trade-offs~\cite{zhang2015deep, li2020federated}, which will be even worse in FL due to the Non-IID data.
Some other work~\cite{wang2018atomo, caldas2018expanding} focus on  reducing the size of the transmitted messages at each round.
These works mainly adopt model-compression schemes such as sparsification and quantization technology to reduce the model size, which have been extensively studied~\cite{wang2018atomo}.
These approaches, however, cannot really speedup model convergence and often lead to accuracy degradation~\cite{yang2020heterogeneity}.
As comparison, our work orchestrates FL protocol with LAN and thus can accelerate model convergence without sacrificing accuracy.

\noindent \textbf{P2P fully decentralized training}~\cite{vanhaesebrouck2017decentralized,lian2017can,colin2016gossip,tang2018d,lalitha2019peer} is an extreme desing point among many learning paradigms where  there is no longer a global state of the model as in standard FL, but the process can be designed such that all local models converge to the desired global solution, i.e., the individual models gradually reach consensus.
Fully decentralized training, however, suffers from slow convergence especially with slow WAN connectivity between geographically-distributed mobile devices.
Our proposal of LAN-aware FL combines both centralized FL and P2P network topology (only within LAN) and thus delivers better convergence performance.
	\section{Discussion}

We now discuss the potential limitations of \sys{}'s current design and future directions to be explored.

\textbf{LAN introduces additional bias in device selection.}
At its current design, \sys{} only orchestrates the LAN domains with certain number of available devices (e.g., at least 10) for each cloud round.
It indicates that devices that are not within such qualified domains will not participate in the training, which potentially introduces additional bias on the global model obtained.
While the bias might exist, we justify our design in two aspects:
(1) Even in traditional FL protocol, a large quantity of devices (more than 30\%) that will never participate in the training~\cite{yang2020heterogeneity}.
(2) FL process typically takes many hours, during which a device has high chance to be connected to a qualified LAN domain.

Given those justifications, we are also seeking to improve \sys{} to embrace dynamic LAN domain sizes so that the domains with only a small number of devices (e.g., only 1 device) can also pariticpate in the training and their results can be properly aggregated towards the global optima.
Our preliminary experiments already show that classical FL protocols, e.g., \textit{FedAvg} do not work for such heterogeneous, hierarchical aggregation.

\textbf{Ad-hoc parameter tuning for global/local aggregation.}
The current design of \sys{} relies on manual parameter tuning through extensive experiments, e.g., the local training epoch, local/global aggregation round number, etc.
While the evaluation demonstrates the effectiveness of our LAN-aware design (goal of this work), such ad-hoc tuning can impose high engineering overhead on FL practitioners.
Instead, we plan to explore automatic techniques~\cite{bonawitz2019federated,kairouz2019advances} to further enhance the usability of \sys.

\textbf{Simulation vs. deployment}
\sys{} is simulation-based as existing FL platforms~\cite{leaf, yang2020heterogeneity, ttf2019, fate2019}.
Though our simulation is based on real-world profiled metrics, e.g., the on-device training speed and communication throughput, some other influential factors might be ignored such as device statues.
Besides, \sys{}'s design in constructing network toplogy ($\S$\ref{subsec:lan-aware-cc}) is evaluated under simple simulation environment, e.g., 1 and 4 APs.
We plan to evaluate \sys{} under more practical settings through real-world deployment. 

\section{Conclusions}
\label{sec:conclusion}

Traditional federated learning protocols heavily rely on wide-area network that often slows down the learning process and adds billing cost to developers.
To address those raised limitations, this work proposes a new federated learning platform \sys, which leverages both local-area network for frequent local aggregation and wide-area network for infrequent global aggregation.
Through such a hierarchical structure, \sys{} can sigificantly accelerate the training process and reduce the wide-area network traffic.
	
	\bibliographystyle{plain}
	\interlinepenalty=10000 

    \bibliography{yuan_ref}
	
	
\end{document}